\documentclass{article}

\usepackage[preprint]{neurips_2026}

\usepackage[utf8]{inputenc}
\usepackage[T1]{fontenc}
\usepackage{hyperref}
\usepackage{url}
\usepackage{microtype}
\usepackage{graphicx}
\usepackage{subcaption}
\usepackage{booktabs}
\usepackage{amsmath}
\usepackage{amssymb}
\usepackage{amsfonts}
\usepackage{mathtools}
\usepackage{amsthm}
\usepackage{tabularx}
\usepackage{multirow}
\usepackage{colortbl}
\usepackage[table,xcdraw]{xcolor}
\usepackage[capitalize,noabbrev]{cleveref}

\theoremstyle{plain}

\theoremstyle{definition}

\theoremstyle{remark}

\title{A Progressive Training Strategy for Embodied Vision-Language Models to Mitigate Spatio-Temporal Hallucinations}
\author{
Xiaoda Yang\thanks{Equal contribution.} \\
Zhejiang University
\And
Shuai Yang\footnotemark[1] \\
The Hong Kong University of Science and Technology
\And
Can Wang \\
Qingdao University
\And
Jingyang Xue \\
Qingdao University
\And
Menglan Tang \\
Qingdao University
\And
Checheng Yu \\
The University of Hong Kong
\And
Xunzhe Zhou \\
The University of Hong Kong
\And
Sashuai Zhou \\
Zhejiang University
\And
Tao Jin \\
Zhejiang University
\And
Lixin Yang \\
Shanghai Jiao Tong University
\And
Xiangyu Yue \\
The Chinese University of Hong Kong
\And
Zhou Zhao\thanks{Corresponding author: zhaozhou@zju.edu.cn} \\
Zhejiang University
}

\begin{document}

\maketitle
\vspace{-0.16in}
\noindent\begin{minipage}{\textwidth}
  \captionsetup{type=figure}
  \centering
  \includegraphics[width=\textwidth]{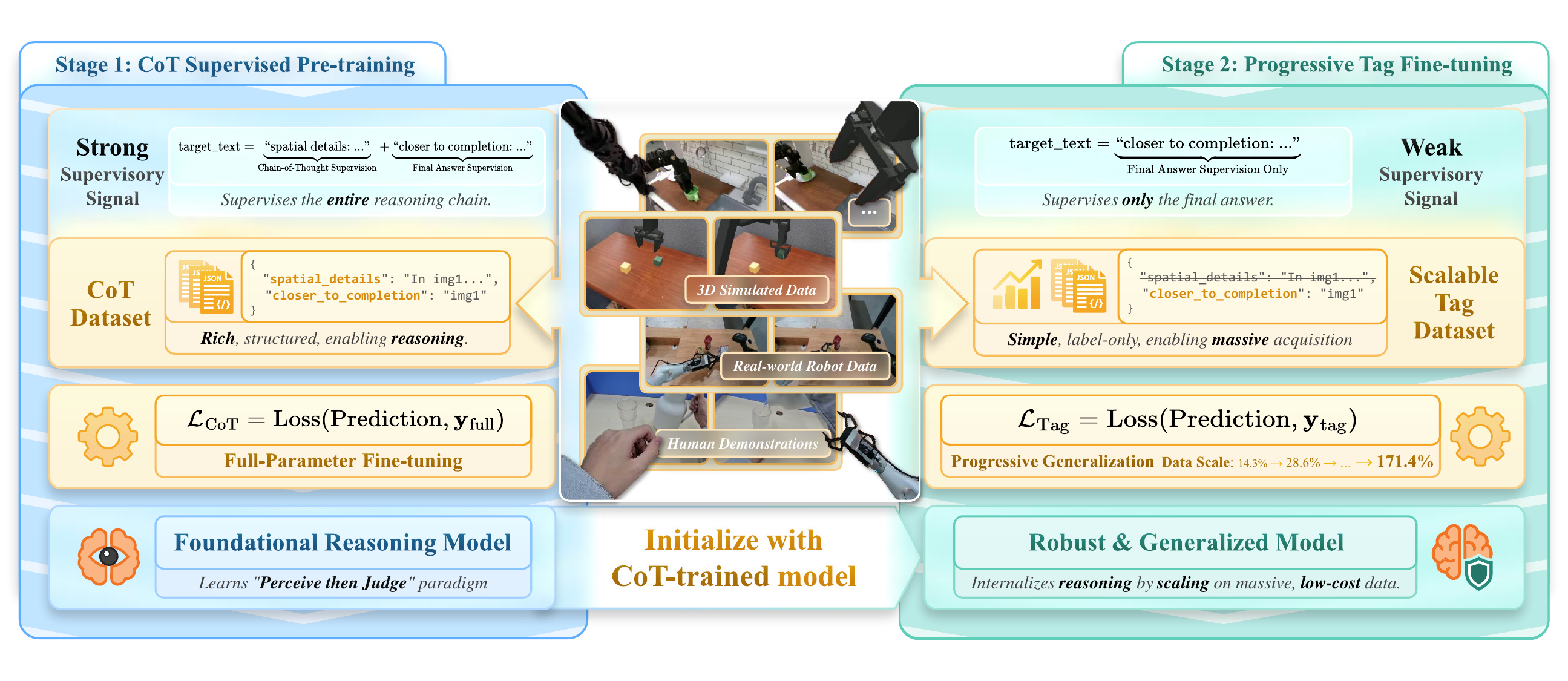}
  \caption{
      Our progressive training paradigm to mitigate spatio-temporal hallucinations. 
      \textbf{Stage 1 (Left):} A CoT-Supervised Pre-training stage instills state-based reasoning by supervising the entire reasoning chain. 
      \textbf{Stage 2 (Right):} A Weakly-Supervised Fine-tuning stage scales this ability using a massive, tag-only dataset, showing a positive scaling trend.
  }
  \label{fig:teaser}
\end{minipage}
\vspace{0.02in}

\begin{abstract}
Vision-Language Models (VLMs) have made significant strides in static image understanding but continue to face critical hurdles in spatiotemporal reasoning. A major bottleneck is ``multi-image reasoning hallucination'', where a large performance drop between forward and reverse temporal queries reveals a dependence on superficial shortcuts instead of state-based understanding. To mitigate this, we first develop a new Chain-of-Thought (CoT) dataset that decomposes intricate reasoning into detailed spatiotemporal steps and definitive judgments. Building on this, we present a progressive training framework: it initiates with supervised pre-training on our CoT dataset to instill logical structures, followed by fine-tuning with scalable weakly-labeled data for broader generalization. Our experiments demonstrate that this approach not only improves backbone accuracy but also reduces the forward-backward performance gap from over $70\%$ to only $6.53\%$. This shows that the method strengthens dynamic reasoning and reduces the inherent temporal biases of current VLMs.
\end{abstract}

\section{Introduction}
\label{sec:introduction}

Vision-Language Models (VLMs) \cite{zhang2024vision} have achieved milestone progress in understanding static images, robustly aligning language with visual perception and laying a solid foundation for a wide range of downstream tasks. However, a critical vulnerability emerges when these models are deployed in dynamic, task-oriented environments: when handling multi-image comparison tasks, they tend to improperly generalize the temporal-logical priors learned during video pre-training. This leads to erroneous judgments of dynamic spatial relationships and hallucinations, particularly in manipulation scenarios. In this context, while formulated as a simple comparison, this task proxies reward modeling by requiring an understanding of physical state transitions rather than mere visual matching.

To address these spatial hallucinations, this work proposes a progressive training framework guided by Chain-of-Thought (CoT) \cite{wei2022chain}. The cornerstone of our approach is a novel, large-scale CoT dataset constructed from videos of daily operational tasks. We first temporally downsample these videos by a factor of ten and then annotate them with detailed reasoning processes. Each data point provides not only the final answer but also a structured analysis of spatial nuances, with this analysis constituting the core supervisory signal. This supervisory approach explicitly teaches the model \textit{how} to reason, thereby establishing a solid foundation for subsequent generalization.

Building on this dataset, we introduce a two-stage, full-parameter fine-tuning paradigm analogous to a student's learning trajectory: first mastering core concepts, then internalizing knowledge through rigorous practice. 
The first stage, \textbf{CoT-Supervised Pre-training}, fully fine-tunes the VLM (Vision-Language Model) on our CoT dataset. 
By supervising the complete reasoning chain (both \texttt{spatial details} and the final \texttt{closer to completion} label), the model is encouraged to adopt a ``perceive-then-judge'' cognitive paradigm, thereby building a robust knowledge framework. 
Subsequently, the curriculum transitions to the second stage: \textbf{Weakly-Supervised Fine-tuning}. 
At this stage, only the final labels are used to fine-tune the model.
Unlike the meticulously crafted CoT data, the data for this stage is easy to acquire, as nearly all operational videos inherently contain the \texttt{closer to completion} supervisory signal, providing a foundation for massive-scale data augmentation. 
By removing the explicit reasoning steps, this stage encourages the model to leverage the foundational cognitive paradigm learned in Stage 1 to arrive at its final judgment. 
This ``study-then-practice'' progression ensures the model grounds its conclusions in visual evidence, ultimately cultivating an internalized, transferable reasoning ability rather than one based on rote memorization.

Our experiments yield strong results. The CoT-only model achieves an average accuracy of 87.07\% on the test set. More importantly, it demonstrates strong robustness against sequence bias, with the performance gap between forward and reverse-ordered tasks being reduced to 6.53\%. This mitigates the severe performance disparity in baseline models, indicating that spatial hallucinations are reduced and the model has transitioned from relying on sequential heuristics to more robust state-based temporal reasoning. Our main contributions are:
\begin{itemize}
    \item \textbf{Large-scale Chain-of-Thought (CoT) dataset:} We construct a novel dataset, STCR-CoT, comprising over 30 million fine-grained spatial annotations derived from real-world operational tasks.
    
    \item \textbf{Low-cost and scalable data engine:} We propose a data generation engine that leverages temporal progress relationships in video. By swapping frame orders and applying multi-scale sampling, it generates massive amounts of weakly-supervised (tag-only) data at a minimal cost while reducing annotation noise through consistency filtering.
    
    \item \textbf{Progressive training paradigm:} We introduce a training paradigm that transitions from strong supervision on our CoT dataset to weak supervision on the scalable, tag-only data. Through this, we empirically observe a positive scaling trend, showing that model performance generally improves as the volume of weakly-supervised data increases.
    
    \item \textbf{Effective mitigation of temporal bias and a novel reward modeling approach:} Through extensive experiments, we demonstrate that our paradigm is effective. It mitigates sequence bias in multi-image spatial reasoning by reducing the forward-reverse performance gap to 6.53\%, and the resulting model can also accurately assess task progression.
\end{itemize}

\section{Related Work}
\label{sec:related_work}

\subsection{Benchmarking Spatio-Temporal Reasoning in Vision-Language Models}
\label{subsec:benchmarking}

Vision-Language Models (VLMs) have evolved from understanding static images to evaluating complex, dynamic scenes. Early benchmarks like CLEVR \cite{johnson2017clevr} focused on compositional geometry in synthetic images, while subsequent work like GQA \cite{hudson2019gqa} extended this to complex real-world imagery.
Focus later shifted towards 3D and dynamic environments. Datasets like ScanNet \cite{dai2017scannet} and Matterport3D \cite{chang2017matterport3d} pioneered 3D understanding, while interactive platforms such as Habitat \cite{savva2019habitat} and AI2-THOR \cite{deitke2022procthor} marked a shift to dynamic tasks requiring navigation and decision-making. Building on this, research emerged to endow VLMs with native 3D perception, leading to models like 3D-LLM, LL3DA, and Chat-Scene. This trend was further advanced by models including 3D-LLaVA and Video-3D LLM, which integrate 3D perception with LLM reasoning. Despite these advances, diagnostic tools like the Perception Test \cite{pavez2023perception} indicate that even state-of-the-art models often lack a deep temporal understanding of events.

\subsection{Task Progress Estimation and Reward Model}
\label{subsec:progress_estimation}

Evaluating task progress is crucial, leading to various methods for constructing general-purpose reward models. One approach learns progress representations via time-contrastive learning, where works like \cite{ma2022vip} leverage the temporal order of video frames as a self-supervised signal. Another approach utilizes the zero-shot capabilities of VLMs, using prompts to score or compare states \cite{yang2024octo}. Furthermore, other studies measure progress via semantic similarity, mapping states and goals to a shared embedding space \cite{mendonca2023goal}, or by synthesizing target states for comparison \cite{zhou2024genie}. These signals frequently provide feedback for Reinforcement Learning \cite{chen2024vision}.

\subsection{Chain-of-Thought for Multimodal Reasoning}
\label{subsec:cot}

Chain-of-Thought (CoT) enhances LLM reasoning by prompting models to generate intermediate steps before the final answer \cite{wei2022chain}. This method effectively handles arithmetic, commonsense, and symbolic reasoning. Recently, CoT has been extended to the multimodal domain for complex visual tasks. Most multimodal CoT applications operate at inference-time using structured prompts to guide generation \cite{zhang2023multimodal,rose2024videocot}.

\section{Dataset}
\label{sec:Dataset}
A key vulnerability of vision-language models (VLMs) is their strong sensitivity to input order during spatio-temporal reasoning. This sensitivity manifests as severe hallucinations, particularly when the logical progression of events contradicts the visual sequence presented to the model. For instance, when tasked to determine which of two images is closer to task completion, merely reversing the input order can cause a large failure in reasoning, a phenomenon termed ``multi-image reasoning hallucination''.
A key factor is supervisory misalignment: while modern VLMs possess the latent capacity for spatial reasoning, standard image-caption training encourages them to rely on superficial temporal shortcuts (e.g., file order) rather than state-change analysis. These paradigms typically rely exclusively on sparse, end-to-end labels. Such weak supervision is insufficient to counteract the strong ``sequence bias'' inherent from large-scale pre-training. Consequently, the model defaults to an erroneous prior: associating task progression with the second image by default, rather than reasoning from visual evidence. To break this vicious cycle, a dataset with a novel supervisory structure is crucial. Our goal is to replace this inadequate sparse supervision with a dense, structured signal that encourages the model to follow a verifiable reasoning process, thereby decoupling its judgment from superficial positional cues.

\subsection{Preprocessing and Sampling}
To construct reliable training samples, our data processing pipeline begins with the refinement of the AgiBot \cite{agibotworldcontributors2025agibotworldcolosseolargescale} real-world robot video corpus. First, we employ a multi-granularity sampling method, applying 10x temporal downsampling to filter out visual noise (such as camera shake) and ensure significant state changes between frames. This process encourages the model to move beyond simple pattern matching. Second, we utilize multi-granularity sliding windows (with intervals of 5, 10, 12 frames, etc.) to extract image pairs. This multi-granularity design creates a natural difficulty curriculum: smaller windows test the model's fine-grained visual discrimination capabilities, whereas larger windows demand high-level logical inference. Building upon this, we perform an important optimization step. To reduce potential data noise and improve training efficiency, we programmatically trim redundant operations from the end of many videos that are irrelevant to the core task (e.g., the robotic arm resetting). This refinement process keeps the selected image pairs focused on task progression, thereby laying a solid foundation for subsequent annotation.

The structured CoT annotations are generated by a dedicated data engine rather than by direct manual labeling at scale, which would be infeasible for 34.7M samples. For each sampled pair, we use a structured prompt to guide Seed1.6-Pro to produce pseudo-labels for fine-grained spatial differences between high-frequency frames, including the intermediate \texttt{spatial details} used by our first-stage supervision. We then apply a bidirectional consistency filter: the same video segment is queried in both forward and reverse order, and samples that lead a strong model to inconsistent judgments are removed. Finally, we randomly sample 10,000 generated CoT examples for cross-human inspection, checking both the final task-progress judgment and the spatial reasoning chain. This quality-control pipeline is used before injecting CoT annotations into Stage 1 training. To reduce train-test coupling, we assign source episodes to a single split before sliding-window expansion; forward/reverse variants and overlapping windows are therefore kept within the same split rather than shared across training and evaluation. Although Seed1.6-Pro is used as the CoT teacher, evaluation is conducted with held-out task-progress labels and includes Seed and non-Seed model families, so the Seed1.6 baseline is used as a reference point rather than as the source of the test labels.

\begin{figure*}[htbp]
    \centering
    \includegraphics[width=\textwidth]{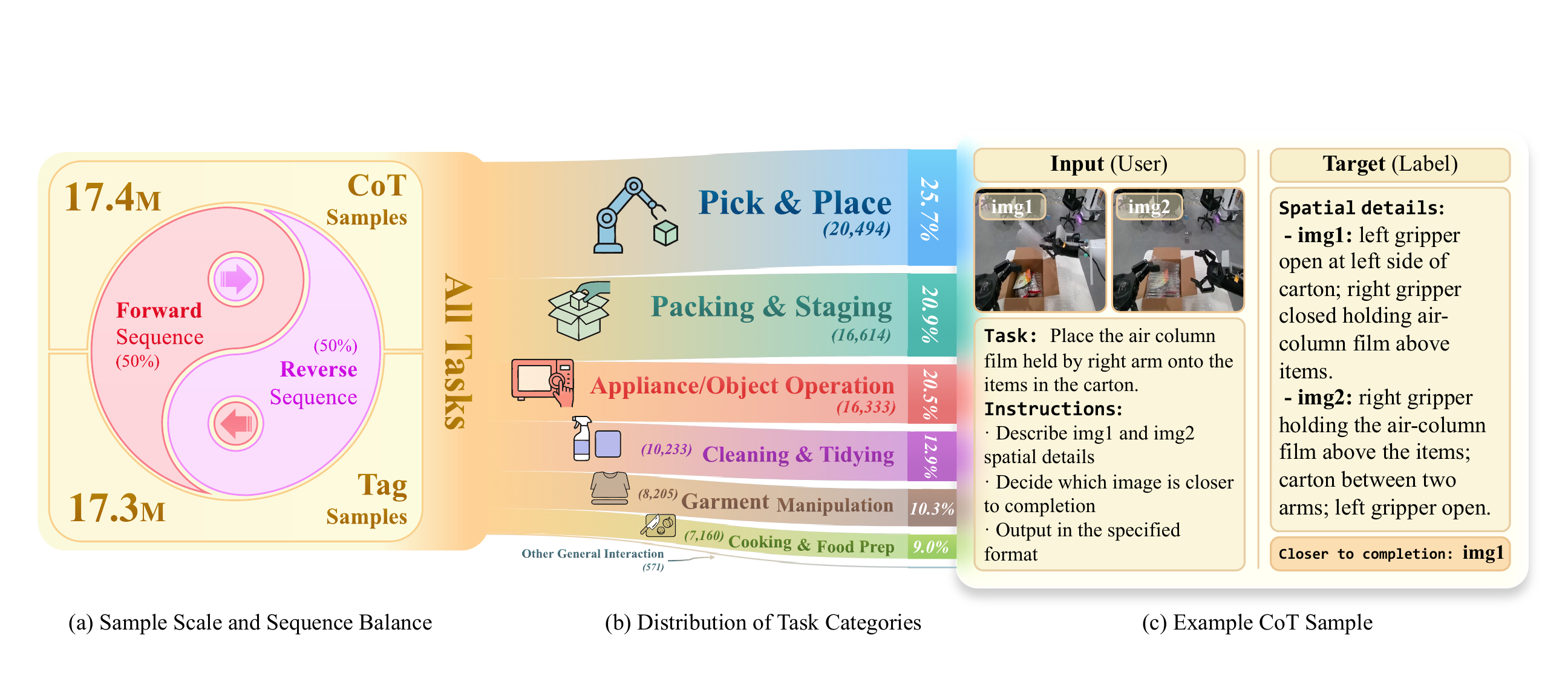}
    \caption{
        \textbf{Statistical analysis of the STCR-CoT dataset.}
        \textbf{(a) Sample Scale and Sequence Balance:} The dataset reaches a total scale of 34.7 million samples and features a globally balanced design, with forward and reverse sequences each constituting 50\% to eliminate temporal bias.
        \textbf{(b) Distribution of Task Categories:} The distribution of task categories is diverse, covering 7 major categories including ``Pick \& Place" and ``Packing \& Staging", across 79,610 unique task instances.
        \textbf{(c) Window Size Distribution:} A progressive learning curriculum is established using a multi-scale sliding window strategy (5-16 frames), with difficulty advancing from fine-grained visual discrimination (smaller windows) to macro-level logical inference (larger windows).
    }
    \label{fig:Statistical}
\end{figure*}

\subsection{Forward and Inverse Contrasting for Decoupling Order Bias}
\label{subsec:forward_inverse_contrasting}

We organize the data by pairing forward and inverse instances of the same example. Since the visual and semantic context is largely held constant, the model must focus on the single variable of input order: a shallow heuristic learned from one sample (e.g., ``select the second image") is immediately contradicted by its inverse counterpart. This counterbalancing discourages reliance on positional priors and repeatedly guides the model to inspect visual state changes and identify a judgment criterion valid under both sequences. In practice, this makes each paired example serve as a local consistency check: the correct answer changes only when the queried order changes, while the underlying evidence still has to be explained by object states, contact relations, and task progress. Through this design, we encourage a more general state-comparison capability rather than shortcuts tailored to a specific ordering pattern.

\subsection{Data Statistics}
The large-scale Spatio-Temporal Causal Reasoning Chain-of-Thought (STCR-CoT) dataset constructed in this study comprises a total of 34.7 million filtered samples. We have implemented a balanced design of forward and reverse samples at a global level. As illustrated in Figure~\ref{fig:Statistical}(a), the dataset ensures that forward and reverse sequences each account for 50\% through meticulous sequence construction. This global balancing design directly targets the issue of temporal bias.

    The dataset is composed of two parts: (1) Chain-of-Thought (CoT) Samples: These provide strong supervision signals. Each sample includes an image pair, a final answer label, and a structured reasoning chain with spatial details. (2) Tag Samples: These offer weak supervision signals. The data is obtained from numerous manipulation videos, with each sample containing only an image pair and its corresponding binary classification label.

The dataset covers 79,610 unique task instances, exhibiting a well-diversified category distribution as shown in Figure~\ref{fig:Statistical}(b): Pick and Place (25.7\%), Packing and Staging (20.9\%), Appliance/Object Manipulation (20.5\%), Cleaning and Tidying (12.9\%), Garment Manipulation (10.3\%), Cooking and Food Preparation (9.0\%), and others.

Furthermore, a progressive difficulty curriculum is constructed using a multi-scale sliding window sampling strategy (5--16 frames), as depicted in Figure~\ref{fig:Statistical}(c). Samples from smaller windows (5--9 frames) focus on fine-grained visual discrimination, while those from larger windows (12--16 frames) emphasize macro-level reasoning. The distribution, where the number of samples increases with the window size, provides a systematic learning path for the model.

In summary, with its substantial scale (34.7M samples), global forward-reverse balancing design, diverse task coverage (7 major categories, 79K unique tasks), and multi-scale curriculum planning, the STCR-CoT dataset provides a strong data foundation for training Vision-Language Models with robust spatio-temporal reasoning capabilities. This holistic balancing design helps the model learn reasoning patterns that are less dependent on the input sequence order.

\section{Method}
\label{sec:method}

In end-to-end training, large models tend to learn superficial statistical shortcuts rather than a structured cognitive process that proceeds from precise perception to logical judgment. As a result, their decision-making lacks a solid evidential foundation and causes ``hallucinations". To address this challenge, we propose a progressive training paradigm based on Chain-of-Thought (CoT). This paradigm emulates the cognitive progression of a student who first intensively studies core knowledge and then engages in extensive practice for deep comprehension and internalization.

\subsection{CoT-Driven Spatial Perception Pre-training: From Shortcut Learning to Structural Grounding}
\label{subsec:cot_pretraining}

The key vulnerability of Vision-Language Models (VLMs) in spatiotemporal reasoning lies in \textbf{Shortcut Learning}. When optimized via sparse end-to-end labels, models tend to converge on ``statistical coincidences''---such as the sequential order of input frames---rather than learning visual state transitions. To address this, Stage 1 of our paradigm treats Chain-of-Thought (CoT) as more than a data format: it serves as a \textbf{Structural Bottleneck} that reshapes the model's optimization trajectory.

\textbf{The ``Perceive-then-Judge'' Cognitive Scaffold.} 
In this phase, we treat the model as a novice required to master a ``core textbook'' of spatial relationships. By extending the supervisory signal from the sparse final judgment to the dense intermediate reasoning chain (the \texttt{spatial details}), we guide the model toward a more structured internal computation. Instead of learning a direct, high-entropy mapping from pixels to answers, the model is encouraged to follow a \textbf{unidirectional information flow} within its computational graph:
\begin{enumerate}
    \item \textbf{The Perception Constraint:} The model must first compress raw visual features into verifiable spatial attributes (e.g., object contact, relative distance, and orientation).
    \item \textbf{The Logical Derivation:} The final judgment must be derived exclusively from these extracted attributes, acting as a secondary inference layer.
\end{enumerate}

This mechanism acts as a \textbf{cognitive scaffold}. We use this term operationally: the training target requires intermediate spatial attributes before the final task-progress judgment, rather than only a direct class label. In terms of the model's training dynamics, this dense supervision encourages the gradients for the ``judgment'' logic to depend on the extracted visual evidence. Consequently, the easier path for loss minimization is less likely to bypass visual perception.

\textbf{Countering the Incentive for Shortcut Discovery.} 
A critical risk in large-scale training is \textbf{Incentive Conflict}: if a model can achieve low training loss by exploiting a brittle correlation (e.g., associating the second image position with task completion), it may ignore complex state comparisons. Our progressive strategy mitigates this by front-loading the ``rigorous reasoning'' objective. By the time the model encounters larger, weakly-labeled datasets in Stage 2, its representation has already been optimized under the perceive-then-judge target. This \textbf{structural initialization} makes the model more resistant to erroneous order priors present in raw video data.

\subsection{Scaling Trend for a Scalable Training Paradigm}
\label{subsec:scaling_laws}

After the model masters fine-grained reasoning ability through pre-training, it transitions to a weakly-supervised fine-tuning stage. This stage simulates a student moving from teacher-guided study to independent, extensive practice. We initialize this stage with the model from the first phase and fine-tune it on a large-scale, diverse database containing only final-answer labels (e.g., labels indicating ``which state is closer to task completion''). By removing supervision on the intermediate steps, we encourage the model to rely on the reasoning pathways learned during pre-training. This entire approach is strategically designed to leverage a crucial advantage for scalability: data requiring only final labels is abundant and easy to acquire. Our experiments reveal a clear positive trend: as the volume of weakly-supervised data increases, model performance generally improves. This finding demonstrates the scaling potential of our training paradigm.

If the CoT pre-training stage is akin to studying a ``textbook with answers and detailed explanations'' to build a robust cognitive ``framework'' that is harder to bypass with shortcuts, then this stage is analogous to practicing with a ``massive problem bank with answers only'' to achieve deeper comprehension. Throughout this process, our progressive fine-tuning strategy plays a guiding role. It encourages the model to apply and consolidate this internal structure in every judgment, thereby efficiently ``distilling'' structured CoT supervision and generalizing it onto the inexpensive, abundant weakly-supervised data.

\section{Experiments}
\label{sec:experiments}
\begin{figure*}[htbp]
    \centering
    \includegraphics[width=\textwidth]{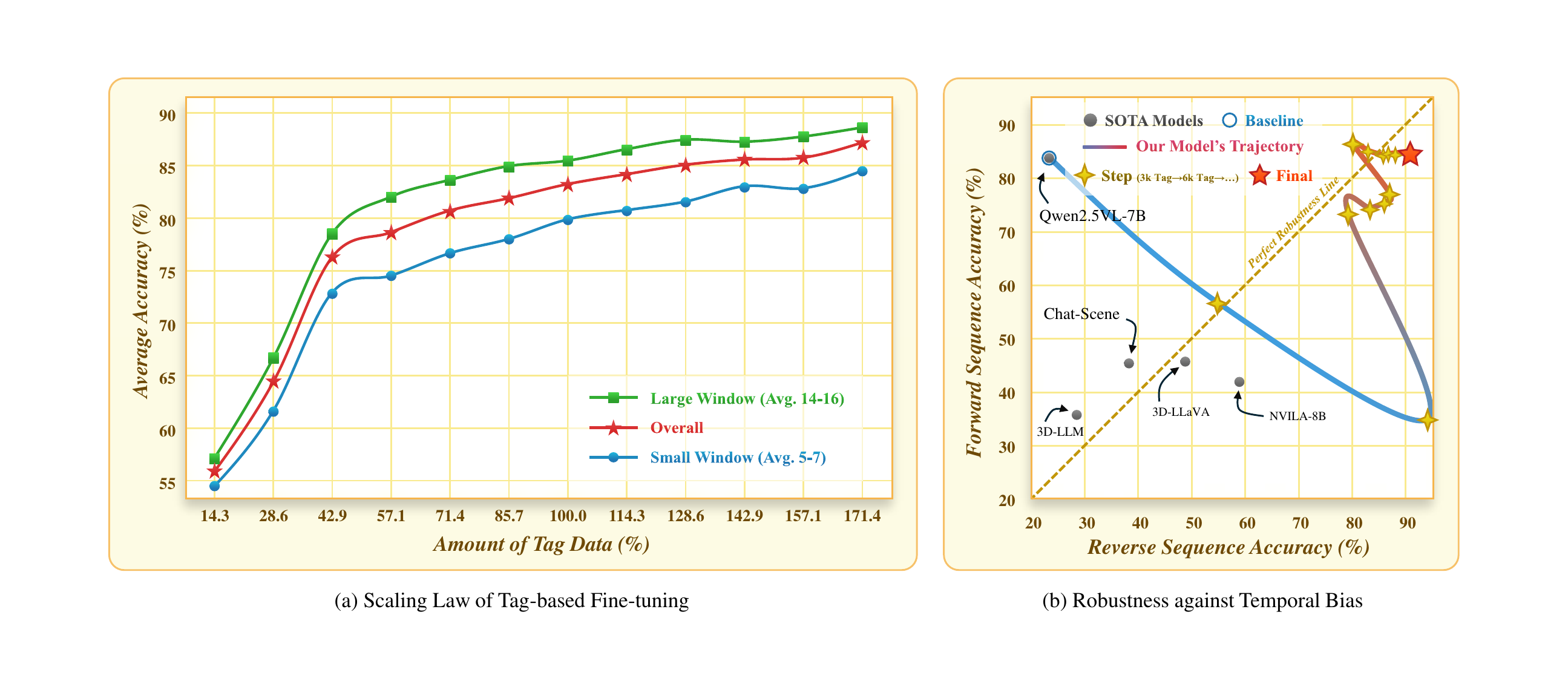}
    \caption{
        \textbf{Main results of our training paradigm.} 
        \textbf{(a) Scaling Trend of Tag-based Fine-tuning:} Model accuracy exhibits a clear positive correlation with the amount of weakly-supervised data. This shows that our paradigm effectively benefits from scaling up in this setting.
        \textbf{(b) Robustness against Temporal Bias:} Our model's trajectory (purple/blue line) demonstrates significantly improved robustness, moving closer to the perfect robustness diagonal compared to the baselines.
    }
    \label{fig:main_results}
\end{figure*}
\subsection{Evaluation metrics}
\label{subsec:evaluation_metrics}

We introduce a large-scale evaluation system for daily-life operations, comprising 35,660 independent test episodes across 29 high-level task categories. The benchmark spans supermarket and retail scenarios (over 7,400 instances, e.g., picking items, packing, and barcode scanning), daily household activities (over 10,000 instances, e.g., dishwashing, ironing, sorting, and folding), kitchen and cooking tasks, and warehousing/logistics operations (over 5,800 instances). These categories cover multi-object interaction, fine-grained manipulation, procedural reasoning, and industrial task-progress judgment, providing a broad testbed for forward/reverse temporal robustness. We also use MMSI-Bench \cite{yang2025mmsi}, a 1,000-question multi-image VQA benchmark with step-by-step reasoning annotations, to evaluate whether the trained model transfers to broader spatial reasoning.

\subsection{Implementation details}
\label{subsec:implementation_details}

In the TAG (label-only) phase, we utilize the TAG weakly supervised dataset, sampled to varying proportions of the total CoT dataset volume ($\left\{ \frac{i}{7} \mid i \in \{1, 2, \dots\} \right\}$). The entire training process uniformly employs the AdamW optimizer \cite{loshchilov2017decoupled}, with a carefully configured learning rate scheduling strategy: we set an initial learning rate of 2e-7 and use a Cosine Learning Rate Scheduler \cite{loshchilov2016sgdr} to smoothly adjust the learning rate. To optimize training efficiency and effectively handle the large-scale model and data, the batch size is set to 8, combined with a 2-step Gradient Accumulation \cite{goyal2017accurate} strategy, which is computationally equivalent to a larger effective batch size. Furthermore, to ensure training stability and accelerate convergence, a 1000-step warmup phase is implemented. To improve training speed and reduce memory footprint, BF16 mixed precision is enabled throughout the entire training process.

Stage 1 uses teacher-forced autoregressive supervision over a fixed response schema consisting of \texttt{spatial details} followed by the final \texttt{closer to completion} answer. Stage 2 removes the rationale tokens and supervises only the final binary progress label. All evaluated models use the same image order, prompt template, and deterministic decoding protocol for forward and reverse queries.

Our main two-stage results are obtained with full-parameter fine-tuning in both Stage 1 and Stage 2. LoRA \cite{hu2021lora} is used only in the preliminary backbone-selection study: we first tune several candidate VLM backbones under comparable small-scale LoRA settings to estimate their potential and optimization behavior, and then select Qwen2.5-VL-7B for the large-scale full-parameter training run. The final model reported in the main experiments is trained on 256 H20 GPUs for approximately one month.

\subsection{Main Results: From Positional Heuristics to State-Based Grounding}
\label{subsec:main_results}

\begin{table*}[t]
  \centering
  \caption{Evaluation results across different WinSize and Models} 
  \vspace{1em}
  \label{tab:evaluation_results_winsize} 
  {\setlength{\tabcolsep}{3pt}%
  \begin{tabularx}{\textwidth}{@{} >{\raggedright\arraybackslash}p{0.35\textwidth} *{8}{>{\centering\arraybackslash}X} @{} } 
    \toprule
    \multirow{2}{*}{\textbf{Models}} & \multicolumn{8}{c}{\textbf{WinSize}} \\
    \cmidrule(lr){2-9} 
    & \textbf{5} & \textbf{6} & \textbf{7} & \textbf{8} & \textbf{9} & \textbf{10} & \textbf{11} & \textbf{$\geq$12} \\ 
    \midrule
    \rowcolor[HTML]{EFEFEF} 
    \multicolumn{9}{l}{\textbf{Human Perception}} \\
    Human Perception & 97.3 & 97.3 & 98.2 & 99.1 & 99.1 & 100.0 & 100.0 & 100.0 \\
    \midrule 
    \rowcolor[HTML]{EFEFEF} 
    \multicolumn{9}{l}{\textbf{Proprietary Models}} \\ 
    GPT-o4mini \cite{openai2025openai} & 70.0 & 70.3 & 72.8 & 67.7 & 60.0 & 67.5 & 67.7 & 70.3 \\
    GPT-o4mini-high \cite{OpenAI2024gpt4o} & 70.5 & 72.7 & 71.8 & 72.3 & 75.5 & 77.3 & 78.1 & 78.6 \\
    GPT-4o \cite{OpenAI2024gpt4o} & 71.1 & 71.7 & 76.7 & 77.8 & 79.2 & 79.2 & 79.2 & 79.2 \\
    GPT-o3 \cite{openai2025openai} & 67.2 & 61.4 & 59.6 & 61.6 & 59.6 & 64.0 & 62.0 & 66.6 \\
    Gemini-2.5-Pro \cite{comanici2025gemini} & 71.3 & 75.0 & 72.8 & 75.0 & 77.2 & 80.9 & 89.7 & 100.0 \\
    Doubao-1.5-pro \cite{team2025doubao} & 64.0 & 76.0 & 56.0 & 46.0 & 56.9 & 86.0 & 90.0 & 96.0 \\
    Seed1.6 \cite{seed2025technical}  & 66.2 & 66.8 & 70.6 & 73.3 & 83.2 & 82.9 & 86.8 & 98.3 \\
    \midrule 
    \rowcolor[HTML]{EFEFEF} 
    \multicolumn{9}{l}{\textbf{Open-Source Models (2D Perception)}} \\ 
    NVILA-8B \cite{liu2024nvila} & 46.5 & 52.0 & 46.0 & 56.1 & 50.0 & 50.0 & 47.5 & 55.0 \\
    Paligemma \cite{zhai2024paligemma} & 43.8 & 45.1 & 44.2 & 45.7 & 42.3 & 43.6 & 42.5 & 46.1 \\
    Qwen2.5-VL-7B \cite{qwen2025qwen25vl} & 53.1 & 55.4 & 55.0 & 57.3 & 55.2 & 56.1 & 56.0 & 56.5 \\
    InternVL-8B \cite{chen2024internvl} & 47.7 & 50.6 & 49.4 & 47.9 & 47.1 & 48.7 & 48.9 & 51.0 \\
    DeepSeek-VL2 \cite{wu2024deepseekvl2mixtureofexpertsvisionlanguagemodels} & 64.4 & 66.9 & 65.2 & 67.8 & 67.0 &69.1 & 77.4 & 81.0 \\
    LLaVA-OneVision-7B \cite{chen2024onevision} & 54.8 & 55.4 & 53.5 & 49.8 & 52.1 & 50.6 & 52.7 & 50.4 \\
    \midrule 
    \rowcolor[HTML]{EFEFEF} 
    \multicolumn{9}{l}{\textbf{Open-Source Models (3D Perception)}} \\ 
    3D-LLM \cite{3dllm} & 27.0 & 28.0 & 30.5 & 30.3 & 33.9 & 32.5 & 36.8 & 37.7 \\ 
    LL3DA \cite{chen2023ll3da} & 36.0 & 33.9 & 32.9 & 34.5 & 36.9 & 33.3 & 39.6 & 36.0 \\
    Chat-Scene \cite{huang2024chat} & 30.8 & 32.5 & 40.7 & 49.0 & 41.6 & 45.3 & 43.3 & 51.0 \\
    3D-LLaVA \cite{deng20253dllava} & 43.2 & 46.1 & 46.4 & 47.9 & 49.6 & 48.8 & 48.0 & 47.9 \\
    Video-3D LLM \cite{zheng2024video3dllmlearningpositionaware} & 50.0 & 50.2 & 50.8 & 51.1 & 49.5 & 50.4 & 51.1 & 49.3 \\
    \midrule 
    \rowcolor[HTML]{EFEFEF} 
    \multicolumn{9}{l}{\textbf{Our models}} \\ 
    STCR-CoT & 68.3 & 70.2 & 72.2  &73.7 & 74.5 & 75.6 & 76.7 & 77.8\\
    STCR-Tag & 82.4 & 84.8 & 86.2 & 87.1 & 87.7 & 88.6 & 88.6 & 88.7\\
    \bottomrule
  \end{tabularx}}
\end{table*}


The empirical results summarized in Table~\ref{tab:evaluation_results_winsize} reveal a large gap between current VLMs and human-level spatiotemporal reasoning. 2D-perception models generally outperform 3D-perception models, suggesting that current 3D representations do not yet translate spatial geometry into high-level task-progress semantics. The positive correlation between WinSize and accuracy further indicates that larger temporal windows provide clearer progress cues, while smaller windows require more fine-grained visual discrimination.

The impact of our paradigm is clearest on Qwen2.5-VL-7B. Stage 1 (STCR-CoT) anchors the model in explicit spatial attributes and already makes it competitive with strong proprietary baselines. Stage 2 (STCR-Tag) then shows a positive scaling trend: once the perceive-then-judge scaffold is established, scalable weak labels can further improve reasoning instead of reinforcing shortcut learning. The zero-shot baselines in Table~\ref{tab:evaluation_results_winsize} are mainly used to diagnose order bias in existing VLMs, while the controlled supervision comparison is given by the same-backbone ablations in Table~\ref{tab:ablation_supervision}.

Beyond absolute accuracy, Figure~\ref{fig:main_results}(b) shows that proprietary models such as GPT-4o collapse under sequence reversal, exposing a strong sequence bias. Our paradigm reduces the forward-reverse gap to 6.53\%, indicating that the model bases its decision on physical state rather than frame order.

\begin{figure*}[htbp]
    \centering
    \includegraphics[width=\textwidth]{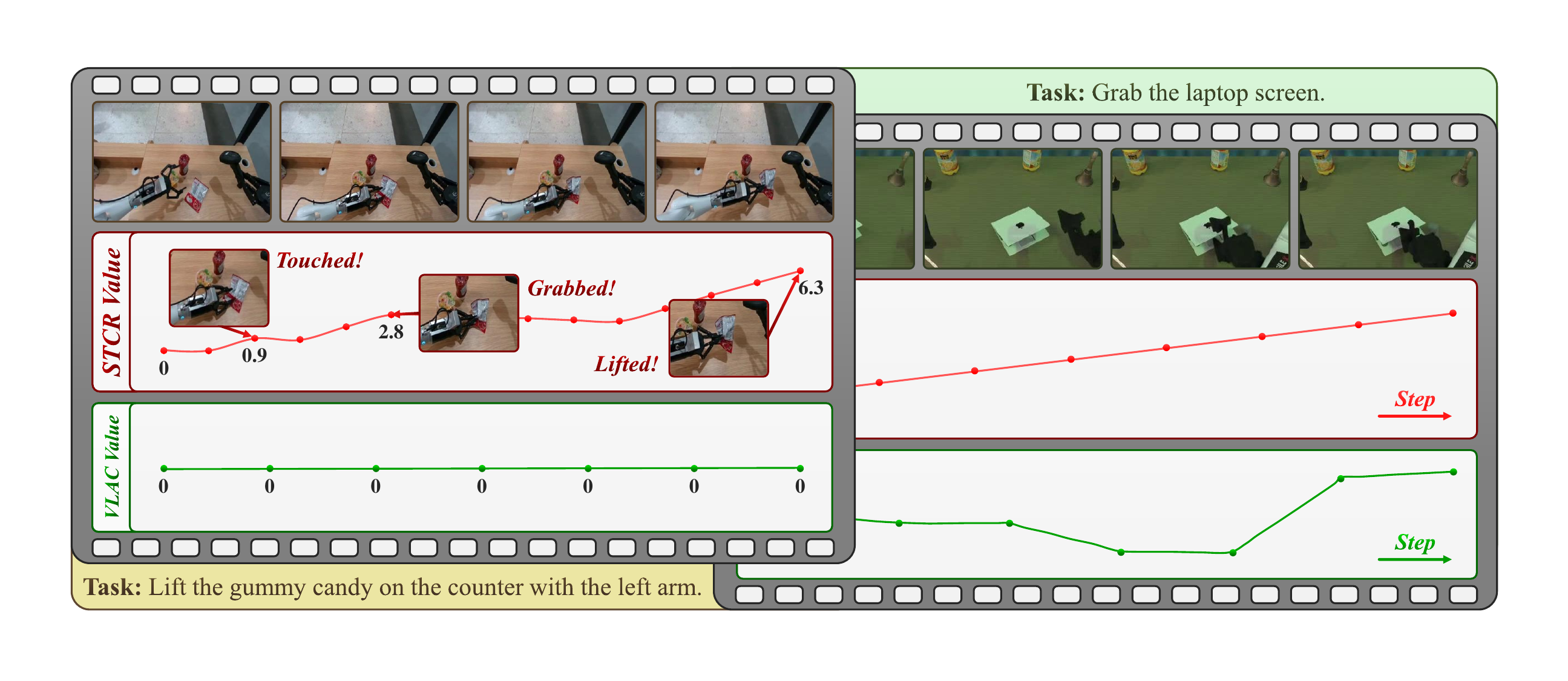} 
    \caption{
        \textbf{Validation of Temporal Consistency in Embodied Multi-Image Reasoning.} 
        The figure compares the reward signals from our model (top) against the baseline VLAC (bottom) \cite{VLAC2025}.
    }
    \label{fig:Comparison}
\end{figure*}

Finally, the reward-model diagnostic in Figure~\ref{fig:Comparison} shows that our model produces a more monotonic task-progress signal than VLAC. This suggests that the learned representation can support temporal credit assignment by mapping physical progress to a continuous reward space.

\subsection{Ablation Study: Dissecting the Structural Bottleneck}
\label{subsec:ablation_study}

To diagnose the learning dynamics, we compare our paradigm with classifier-based supervision variants. Table~\ref{tab:ablation_supervision} shows that the supervision signal strongly affects accuracy and robustness.

\begin{table}[htbp]
    \centering
    \caption{Ablation study on supervision strategies}
    \label{tab:ablation_supervision}
{\small\setlength{\tabcolsep}{3pt}%
\begin{tabularx}{\columnwidth}{@{} >{\raggedright\arraybackslash}X >{\centering\arraybackslash}p{0.18\columnwidth} >{\centering\arraybackslash}p{0.18\columnwidth} >{\centering\arraybackslash}p{0.18\columnwidth} @{} }
        \toprule
        \textbf{Strategy} & \textbf{Forward (\%)} & \textbf{Reverse (\%)} & \textbf{Average (\%)} \\
        \midrule
        Classifier-head only          & 80.30 & 9.92  & 45.11 \\
        CoT + Classifier-head & 84.43 & 16.34 & 50.39 \\
        \midrule
        \textbf{CoT-only (Ours)} & \textbf{83.80} & \textbf{90.33} & \textbf{87.07} \\
        \bottomrule
    \end{tabularx}}
\end{table}

\textbf{Failure of End-to-End Classification.} 
The traditional ``Classifier-head only'' strategy suffers a catastrophic \textbf{Symmetry Collapse}, with a 70.38\% forward-reverse gap. This indicates shortcut learning: with only sparse final labels, the model exploits input order instead of grounding the decision in visual evidence.

\textbf{Gradient Interference in Mixed Supervision.} 
``Mixed supervision'' also fails to match CoT-only robustness. Gradients from granular CoT and sparse labels interfere, so the model treats reasoning and classification as parallel targets rather than a single perceive-then-judge hierarchy.

\textbf{The Power of a Generative Bottleneck.} 
In contrast, ``CoT-only'' exclusively supervises the reasoning chain, enforcing a unidirectional information flow from perception to judgment. It achieves the best average accuracy (87.07\%) and reduces the temporal gap to 6.53\%.

\section{Conclusion}
We present a CoT-based progressive training paradigm for improving spatiotemporal reasoning in VLMs. The results show that temporal hallucination is not only a data-scale problem, but also a supervision-structure problem: sparse labels let models exploit input order, while reasoning-chain supervision encourages them to connect visual state changes to task progress. Our method reaches 87.07\% accuracy and reduces the forward-reverse gap to 6.53\%, and external benchmarks and cross-backbone results suggest that this structured intermediate supervision transfers beyond the main evaluation setting.

\section{Limitations}
Our main experiments use Qwen2.5-VL-7B; Appendix~\ref{app:cross_backbone} reports less-tuned InternVL3-8B results. Appendix~\ref{app:external_benchmarks} shows mixed transfer on external benchmarks, and general 3D measurement remains harder. Residual noise and teacher-bias inheritance may remain in generated CoT annotations despite filtering and human cross-checking. We will release code and data artifacts after anonymity constraints are cleared.

\textbf{Impact Statement.} This work may improve progress estimation and robot-learning feedback, but incorrect progress judgments can still misguide downstream policies.

\bibliographystyle{plainnat}
\bibliography{paper}

@InProceedings{johnson2017clevr,
author = {Johnson, Justin and Hariharan, Bharath and van der Maaten, Laurens and Fei-Fei, Li and Lawrence Zitnick, C. and Girshick, Ross},
title = {CLEVR: A Diagnostic Dataset for Compositional Language and Elementary Visual Reasoning},
booktitle = {Proceedings of the IEEE Conference on Computer Vision and Pattern Recognition (CVPR)},
month = {July},
year = {2017}
}

@INPROCEEDINGS{hudson2019gqa,
  author={Hudson, Drew A. and Manning, Christopher D.},
  booktitle={2019 IEEE/CVF Conference on Computer Vision and Pattern Recognition (CVPR)}, 
  title={GQA: A New Dataset for Real-World Visual Reasoning and Compositional Question Answering}, 
  year={2019},
  volume={},
  number={},
  pages={6693-6702},
  doi={10.1109/CVPR.2019.00686}}

@INPROCEEDINGS{dai2017scannet,
  author={Dai, Angela and Chang, Angel X. and Savva, Manolis and Halber, Maciej and Funkhouser, Thomas and Nießner, Matthias},
  booktitle={2017 IEEE Conference on Computer Vision and Pattern Recognition (CVPR)}, 
  title={ScanNet: Richly-Annotated 3D Reconstructions of Indoor Scenes}, 
  year={2017},
  volume={},
  number={},
  pages={2432-2443},
  doi={10.1109/CVPR.2017.261}}

@misc{chang2017matterport3d,
      title={Matterport3D: Learning from RGB-D Data in Indoor Environments}, 
      author={Angel Chang and Angela Dai and Thomas Funkhouser and Maciej Halber and Matthias Nießner and Manolis Savva and Shuran Song and Andy Zeng and Yinda Zhang},
      year={2017},
      eprint={1709.06158},
      archivePrefix={arXiv},
      primaryClass={cs.CV},
      url={https://arxiv.org/abs/1709.06158}, 
}

@inproceedings{savva2019habitat,
  title={{Habitat: A Platform for Embodied AI Research}},
  author={Savva, Manolis and Kadian, Abhishek and Maksymets, Oleksandr and Zhao, Yili and Wijmans, Erik and Jain, Bhavana and Straub, Julian and Liu, Jia and Koltun, Vladlen and Malik, Jitendra and Parikh, Devi and Batra, Dhruv},
  booktitle={Proceedings of the IEEE/CVF International Conference on Computer Vision (ICCV)},
  year={2019}
}

@article{deitke2022procthor,
  title={Ai2-thor: An interactive 3d environment for visual ai},
  author={Kolve, Eric and Mottaghi, Roozbeh and Han, Winson and VanderBilt, Eli and Weihs, Luca and Herrasti, Alvaro and Deitke, Matt and Ehsani, Kiana and Gordon, Daniel and Zhu, Yuke and others},
  journal={arXiv preprint arXiv:1712.05474},
  year={2017}
}

@inproceedings{pavez2023perception,
  author = {Patraucean, Viorica and others},
  title = {Perception Test: A Diagnostic Benchmark for Multimodal Video Models},
  booktitle = {Advances in Neural Information Processing Systems},
  year = {2023},
  volume = {36},
  pages = {42748--42761}
}

@misc{yang2024octo,
      title={Octo: An Open-Source Generalist Robot Policy}, 
      author={Octo Model Team and Dibya Ghosh and Homer Walke and Karl Pertsch and Kevin Black and Oier Mees and Sudeep Dasari and Joey Hejna and Tobias Kreiman and Charles Xu and Jianlan Luo and You Liang Tan and Lawrence Yunliang Chen and Pannag Sanketi and Quan Vuong and Ted Xiao and Dorsa Sadigh and Chelsea Finn and Sergey Levine},
      year={2024},
      eprint={2405.12213},
      archivePrefix={arXiv},
      primaryClass={cs.RO},
      url={https://arxiv.org/abs/2405.12213}, 
}

@misc{mendonca2023goal,
      title={Goal-Conditioned Imitation Learning using Score-based Diffusion Policies}, 
      author={Moritz Reuss and Maximilian Li and Xiaogang Jia and Rudolf Lioutikov},
      year={2023},
      eprint={2304.02532},
      archivePrefix={arXiv},
      primaryClass={cs.LG},
      url={https://arxiv.org/abs/2304.02532}, 
}

@inproceedings{
zhou2024genie,
title={Genie: Generative Interactive Environments},
author={Jake Bruce and Michael D Dennis and Ashley Edwards and Jack Parker-Holder and Yuge Shi and Edward Hughes and Matthew Lai and Aditi Mavalankar and Richie Steigerwald and Chris Apps and Yusuf Aytar and Sarah Maria Elisabeth Bechtle and Feryal Behbahani and Stephanie C.Y. Chan and Nicolas Heess and Lucy Gonzalez and Simon Osindero and Sherjil Ozair and Scott Reed and Jingwei Zhang and Konrad Zolna and Jeff Clune and Nando de Freitas and Satinder Singh and Tim Rockt{\"a}schel},
booktitle={Forty-first International Conference on Machine Learning},
year={2024},
url={https://openreview.net/forum?id=bJbSbJskOS}
}

@misc{chen2024vision,
      title={Vision-Language Models as a Source of Rewards}, 
      author={Kate Baumli and Satinder Baveja and Feryal Behbahani and Harris Chan and Gheorghe Comanici and Sebastian Flennerhag and Maxime Gazeau and Kristian Holsheimer and Dan Horgan and Michael Laskin and Clare Lyle and Hussain Masoom and Kay McKinney and Volodymyr Mnih and Alexander Neitz and Dmitry Nikulin and Fabio Pardo and Jack Parker-Holder and John Quan and Tim Rocktäschel and Himanshu Sahni and Tom Schaul and Yannick Schroecker and Stephen Spencer and Richie Steigerwald and Luyu Wang and Lei Zhang},
      year={2024},
      eprint={2312.09187},
      archivePrefix={arXiv},
      primaryClass={cs.LG},
      url={https://arxiv.org/abs/2312.09187}, 
}

@misc{zhang2023multimodal,
      title={Multimodal Chain-of-Thought Reasoning in Language Models}, 
      author={Zhuosheng Zhang and Aston Zhang and Mu Li and Hai Zhao and George Karypis and Alex Smola},
      year={2024},
      eprint={2302.00923},
      archivePrefix={arXiv},
      primaryClass={cs.CL},
      url={https://arxiv.org/abs/2302.00923}, 
}

@misc{rose2024videocot,
      title={Visual Chain of Thought: Bridging Logical Gaps with Multimodal Infillings}, 
      author={Daniel Rose and Vaishnavi Himakunthala and Andy Ouyang and Ryan He and Alex Mei and Yujie Lu and Michael Saxon and Chinmay Sonar and Diba Mirza and William Yang Wang},
      year={2024},
      eprint={2305.02317},
      archivePrefix={arXiv},
      primaryClass={cs.CL},
      url={https://arxiv.org/abs/2305.02317}, 
}

@inproceedings{wei2022chain,
  title={Chain-of-Thought Prompting Elicits Reasoning in Large Language Models},
  author={Wei, Jason and Wang, Xuezhi and Schuurmans, Dale and Bosma, Maarten and Ichter, Brian and Xia, Fei and Chi, Ed and Le, Quoc V. and Zhou, Denny},
  booktitle={Advances in Neural Information Processing Systems},
  volume={35},
  pages={24824--24837},
  year={2022},
  url={https://proceedings.neurips.cc/paper_files/paper/2022/hash/9d5609613524ecf4f15af0f7b31abca4-Abstract-Conference.html}
}

@misc{qwen2025qwen25vl,
      title={Qwen2.5-VL Technical Report}, 
      author={Shuai Bai and Keqin Chen and Xuejing Liu and Jialin Wang and Wenbin Ge and Sibo Song and Kai Dang and Peng Wang and Shijie Wang and Jun Tang and Humen Zhong and Yuanzhi Zhu and Mingkun Yang and Zhaohai Li and Jianqiang Wan and Pengfei Wang and Wei Ding and Zheren Fu and Yiheng Xu and Jiabo Ye and Xi Zhang and Tianbao Xie and Zesen Cheng and Hang Zhang and Zhibo Yang and Haiyang Xu and Junyang Lin},
      year={2025},
      eprint={2502.13923},
      archivePrefix={arXiv},
      primaryClass={cs.CV},
      url={https://arxiv.org/abs/2502.13923}, 
}

@ARTICLE{zhang2024vision,
  author={Zhang, Jingyi and Huang, Jiaxing and Jin, Sheng and Lu, Shijian},
  journal={IEEE Transactions on Pattern Analysis and Machine Intelligence}, 
  title={Vision-Language Models for Vision Tasks: A Survey}, 
  year={2024},
  volume={46},
  number={8},
  pages={5625-5644},
  doi={10.1109/TPAMI.2024.3369699}}

@inproceedings{yang2025mmsi,
 author = {Yang, Sihan and Xu, Runsen and Xie, Yiman and Yang, Sizhe and Li, Mo and Lin, Jingli and Zhu, Chenming and Chen, Xiaochen and Duan, Haodong and Yue, Xiangyu and Lin, Dahua and Wang, Tai and Pang, Jiangmiao},
 booktitle = {International Conference on Learning Representations},
 editor = {C. Vondrick and B. Hariharan and C. Raffel and L. Pinto and D. Yang and A. Faust},
 pages = {157051--157088},
 title = {MMSI-Bench: A Benchmark for Multi-Image Spatial Intelligence},
 url = {https://proceedings.iclr.cc/paper_files/paper/2026/file/fe5d4bd3e2af823701b5d8f70a3c7602-Paper-Conference.pdf},
 volume = {2026},
 year = {2026}
}

@misc{OpenAI2024gpt4o,
  title={GPT-4o System Card},
  author={{OpenAI} and others},
  year={2024},
  eprint={2410.21276},
  archivePrefix={arXiv},
  primaryClass={cs.CL},
  url={https://arxiv.org/abs/2410.21276},
}

@InProceedings{Chen2024internvl,
    author    = {Chen, Zhe and Wu, Jiannan and Wang, Wenhai and Su, Weijie and Chen, Guo and Xing, Sen and Zhong, Muyan and Zhang, Qinglong and Zhu, Xizhou and Lu, Lewei and Li, Bin and Luo, Ping and Lu, Tong and Qiao, Yu and Dai, Jifeng},
    title     = {InternVL: Scaling up Vision Foundation Models and Aligning for Generic Visual-Linguistic Tasks},
    booktitle = {Proceedings of the IEEE/CVF Conference on Computer Vision and Pattern Recognition (CVPR)},
    month     = {June},
    year      = {2024},
    pages     = {24185-24198}
}

@inproceedings{3dllm,
 author = {Hong, Yining and Zhen, Haoyu and Chen, Peihao and Zheng, Shuhong and Du, Yilun and Chen, Zhenfang and Gan, Chuang},
 booktitle = {Advances in Neural Information Processing Systems},
 doi = {10.52202/075280-0900},
 editor = {A. Oh and T. Naumann and A. Globerson and K. Saenko and M. Hardt and S. Levine},
 pages = {20482--20494},
 publisher = {Curran Associates, Inc.},
 title = {3D-LLM: Injecting the 3D World into Large Language Models},
 url = {https://proceedings.neurips.cc/paper_files/paper/2023/file/413885e70482b95dcbeeddc1daf39177-Paper-Conference.pdf},
 volume = {36},
 year = {2023}
}

@INPROCEEDINGS{chen2023ll3da,
  author={Chen, Sijin and Chen, Xin and Zhang, Chi and Li, Mingsheng and Yu, Gang and Fei, Hao and Zhu, Hongyuan and Fan, Jiayuan and Chen, Tao},
  booktitle={2024 IEEE/CVF Conference on Computer Vision and Pattern Recognition (CVPR)}, 
  title={LL3DA: Visual Interactive Instruction Tuning for Omni-3D Understanding, Reasoning, and Planning}, 
  year={2024},
  volume={},
  number={},
  pages={26418-26428},
  doi={10.1109/CVPR52733.2024.02496}}

@inproceedings{huang2024chat,
 author = {Huang, Haifeng and Chen, Yilun and Wang, Zehan and Huang, Rongjie and Xu, Runsen and Wang, Tai and Liu, Luping and Cheng, Xize and Zhao, Yang and Pang, Jiangmiao and Zhao, Zhou},
 booktitle = {Advances in Neural Information Processing Systems},
 doi = {10.52202/079017-3620},
 editor = {A. Globerson and L. Mackey and D. Belgrave and A. Fan and U. Paquet and J. Tomczak and C. Zhang},
 pages = {113991--114017},
 publisher = {Curran Associates, Inc.},
 title = {Chat-Scene: Bridging 3D Scene and Large Language Models with Object Identifiers},
 url = {https://proceedings.neurips.cc/paper_files/paper/2024/file/cebbd24f1e50bcb63d015611fe0fe767-Paper-Conference.pdf},
 volume = {37},
 year = {2024}
}

@INPROCEEDINGS{deng20253dllava,
  author={Deng, Jiajun and He, Tianyu and Jiang, Li and Wang, Tianyu and Dayoub, Feras and Reid, Ian},
  booktitle={2025 IEEE/CVF Conference on Computer Vision and Pattern Recognition (CVPR)}, 
  title={3D-LLaVA: Towards Generalist 3D LMMs with Omni Superpoint Transformer}, 
  year={2025},
  volume={},
  number={},
  pages={3772-3782},
  doi={10.1109/CVPR52734.2025.00357}}

@INPROCEEDINGS{zheng2024video3dllmlearningpositionaware,
  author={Zheng, Duo and Huang, Shijia and Wang, Liwei},
  booktitle={2025 IEEE/CVF Conference on Computer Vision and Pattern Recognition (CVPR)}, 
  title={Video-3D LLM: Learning Position-Aware Video Representation for 3D Scene Understanding}, 
  year={2025},
  volume={},
  number={},
  pages={8995-9006},
  doi={10.1109/CVPR52734.2025.00841}}

@INPROCEEDINGS{liu2024nvila,
  author={Liu, Zhijian and Zhu, Ligeng and Shi, Baifeng and Zhang, Zhuoyang and Lou, Yuming and Yang, Shang and Xi, Haocheng and Cao, Shiyi and Gu, Yuxian and Li, Dacheng and Li, Xiuyu and Tang, Haotian and Fang, Yunhao and Chen, Yukang and Hsieh, Cheng-Yu and Huang, De-An and Cheng, An-Chieh and Hu, Jinyi and Liu, Sifei and Krishna, Ranjay and Molchanov, Pavlo and Kautz, Jan and Yin, Hongxu and Han, Song and Lu, Yao},
  booktitle={2025 IEEE/CVF Conference on Computer Vision and Pattern Recognition (CVPR)}, 
  title={NVILA: Efficient Frontier Visual Language Models}, 
  year={2025},
  volume={},
  number={},
  pages={4122-4134},
  doi={10.1109/CVPR52734.2025.00390}}

@article{comanici2025gemini,
  title={Gemini 2.5: Pushing the frontier with advanced reasoning, multimodality, long context, and next generation agentic capabilities},
  author={Comanici, Gheorghe and Bieber, Eric and Schaekermann, Mike and Pasupat, Ice and Sachdeva, Noveen and Dhillon, Inderjit and Blistein, Marcel and Ram, Ori and Zhang, Dan and Rosen, Evan and others},
  journal={arXiv preprint arXiv:2507.06261},
  year={2025}
}

@misc{agibotworldcontributors2025agibotworldcolosseolargescale,
      title={AgiBot World Colosseo: A Large-scale Manipulation Platform for Scalable and Intelligent Embodied Systems}, 
      author={AgiBot-World-Contributors and Qingwen Bu and Jisong Cai and Li Chen and Xiuqi Cui and Yan Ding and Siyuan Feng and Shenyuan Gao and Xindong He and Xuan Hu and Xu Huang and Shu Jiang and Yuxin Jiang and Cheng Jing and Hongyang Li and Jialu Li and Chiming Liu and Yi Liu and Yuxiang Lu and Jianlan Luo and Ping Luo and Yao Mu and Yuehan Niu and Yixuan Pan and Jiangmiao Pang and Yu Qiao and Guanghui Ren and Cheng Ruan and Jiaqi Shan and Yongjian Shen and Chengshi Shi and Mingkang Shi and Modi Shi and Chonghao Sima and Jianheng Song and Huijie Wang and Wenhao Wang and Dafeng Wei and Chengen Xie and Guo Xu and Junchi Yan and Cunbiao Yang and Lei Yang and Shukai Yang and Maoqing Yao and Jia Zeng and Chi Zhang and Qinglin Zhang and Bin Zhao and Chengyue Zhao and Jiaqi Zhao and Jianchao Zhu},
      year={2025},
      eprint={2503.06669},
      archivePrefix={arXiv},
      primaryClass={cs.RO},
      url={https://arxiv.org/abs/2503.06669}, 
}

@misc{chen2024onevision,
      title={LLaVA-OneVision: Easy Visual Task Transfer}, 
      author={Bo Li and Yuanhan Zhang and Dong Guo and Renrui Zhang and Feng Li and Hao Zhang and Kaichen Zhang and Peiyuan Zhang and Yanwei Li and Ziwei Liu and Chunyuan Li},
      year={2024},
      eprint={2408.03326},
      archivePrefix={arXiv},
      primaryClass={cs.CV},
      url={https://arxiv.org/abs/2408.03326}, 
}

@misc{zhai2024paligemma,
      title={PaliGemma: A versatile 3B VLM for transfer}, 
      author={Lucas Beyer and Andreas Steiner and André Susano Pinto and Alexander Kolesnikov and Xiao Wang and Daniel Salz and Maxim Neumann and Ibrahim Alabdulmohsin and Michael Tschannen and Emanuele Bugliarello and Thomas Unterthiner and Daniel Keysers and Skanda Koppula and Fangyu Liu and Adam Grycner and Alexey Gritsenko and Neil Houlsby and Manoj Kumar and Keran Rong and Julian Eisenschlos and Rishabh Kabra and Matthias Bauer and Matko Bošnjak and Xi Chen and Matthias Minderer and Paul Voigtlaender and Ioana Bica and Ivana Balazevic and Joan Puigcerver and Pinelopi Papalampidi and Olivier Henaff and Xi Xiong and Radu Soricut and Jeremiah Harmsen and Xiaohua Zhai},
      year={2024},
      eprint={2407.07726},
      archivePrefix={arXiv},
      primaryClass={cs.CV},
      url={https://arxiv.org/abs/2407.07726}, 
}

@misc{wu2024deepseekvl2mixtureofexpertsvisionlanguagemodels,
      title={DeepSeek-VL2: Mixture-of-Experts Vision-Language Models for Advanced Multimodal Understanding}, 
      author={Zhiyu Wu and Xiaokang Chen and Zizheng Pan and Xingchao Liu and Wen Liu and Damai Dai and Huazuo Gao and Yiyang Ma and Chengyue Wu and Bingxuan Wang and Zhenda Xie and Yu Wu and Kai Hu and Jiawei Wang and Yaofeng Sun and Yukun Li and Yishi Piao and Kang Guan and Aixin Liu and Xin Xie and Yuxiang You and Kai Dong and Xingkai Yu and Haowei Zhang and Liang Zhao and Yisong Wang and Chong Ruan},
      year={2024},
      eprint={2412.10302},
      archivePrefix={arXiv},
      primaryClass={cs.CV},
      url={https://arxiv.org/abs/2412.10302}, 
}

@misc{loshchilov2017decoupled,
      title={Decoupled Weight Decay Regularization}, 
      author={Ilya Loshchilov and Frank Hutter},
      year={2019},
      eprint={1711.05101},
      archivePrefix={arXiv},
      primaryClass={cs.LG},
      url={https://arxiv.org/abs/1711.05101}, 
}

@misc{loshchilov2016sgdr,
      title={SGDR: Stochastic Gradient Descent with Warm Restarts}, 
      author={Ilya Loshchilov and Frank Hutter},
      year={2017},
      eprint={1608.03983},
      archivePrefix={arXiv},
      primaryClass={cs.LG},
      url={https://arxiv.org/abs/1608.03983}, 
}

@misc{goyal2017accurate,
      title={Accurate, Large Minibatch SGD: Training ImageNet in 1 Hour}, 
      author={Priya Goyal and Piotr Dollár and Ross Girshick and Pieter Noordhuis and Lukasz Wesolowski and Aapo Kyrola and Andrew Tulloch and Yangqing Jia and Kaiming He},
      year={2018},
      eprint={1706.02677},
      archivePrefix={arXiv},
      primaryClass={cs.CV},
      url={https://arxiv.org/abs/1706.02677}, 
}

@misc{VLAC2025,
      title={A Vision-Language-Action-Critic Model for Robotic Real-World Reinforcement Learning}, 
      author={Shaopeng Zhai and Qi Zhang and Tianyi Zhang and Fuxian Huang and Haoran Zhang and Ming Zhou and Shengzhe Zhang and Litao Liu and Sixu Lin and Jiangmiao Pang},
      year={2025},
      eprint={2509.15937},
      archivePrefix={arXiv},
      primaryClass={cs.RO},
      url={https://arxiv.org/abs/2509.15937}, 
}

@misc{seed2025technical,
  title={Technical Introduction to the Seed1. 6 Model Series},
  author={Seed, ByteDance},
  year={2025}
}

@article{openai2025openai,
  title={OpenAI o3 and o4-mini system card},
  author={OpenAI},
  journal={https://openai. com/index/o3-o4-mini-system-card/},
  year={2025}
}

@misc{team2025doubao,
  title={Doubao-1.5-pro: A High-Performance Sparse MoE Large Language Model},
  author={Team, Doubao},
  year={2025},
  publisher={Accessed}
}

@inproceedings{jia2026omnispatial,
  title={Omnispatial: Towards comprehensive spatial reasoning benchmark for vision language models},
  author={Jia, Mengdi and Qi, Zekun and Zhang, Shaochen and Zhang, Wenyao and Yu, Xinqiang and He, Jiawei and Wang, He and Yi, Li},
  booktitle={International Conference on Learning Representations},
  volume={2026},
  pages={35634--35670},
  year={2026}
}

@inproceedings{yang2025thinking,
  title={Thinking in space: How multimodal large language models see, remember, and recall spaces},
  author={Yang, Jihan and Yang, Shusheng and Gupta, Anjali W and Han, Rilyn and Fei-Fei, Li and Xie, Saining},
  booktitle={2025 IEEE/CVF Conference on Computer Vision and Pattern Recognition (CVPR)},
  pages={10632--10643},
  year={2025},
  organization={IEEE}
}

@article{tong2024cambrian,
  title={Cambrian-1: A fully open, vision-centric exploration of multimodal llms},
  author={Tong, Shengbang and Brown, Ellis and Wu, Penghao and Woo, Sanghyun and Middepogu, Manoj and Akula, Sai C and Yang, Jihan and Yang, Shusheng and Iyer, Adithya and Pan, Xichen and others},
  journal={Advances in Neural Information Processing Systems},
  volume={37},
  pages={87310--87356},
  year={2024}
}

@article{schulman2017proximal,
  title={Proximal policy optimization algorithms},
  author={Schulman, John and Wolski, Filip and Dhariwal, Prafulla and Radford, Alec and Klimov, Oleg},
  journal={arXiv preprint arXiv:1707.06347},
  year={2017}
}

@article{hu2021lora,
  title={Lora: Low-rank adaptation of large language models},
  author={Hu, Edward J and Shen, Yelong and Wallis, Phillip and Allen-Zhu, Zeyuan and Li, Yuanzhi and Wang, Shean and Wang, Lu and Chen, Weizhu},
  journal={arXiv preprint arXiv:2106.09685},
  year={2021}
}

@article{ma2022vip,
  title={Vip: Towards universal visual reward and representation via value-implicit pre-training},
  author={Ma, Yecheng Jason and Sodhani, Shagun and Jayaraman, Dinesh and Bastani, Osbert and Kumar, Vikash and Zhang, Amy},
  journal={arXiv preprint arXiv:2210.00030},
  year={2022}
}

\clearpage     
\appendix  
\section{Additional External Benchmark Results}
\label{app:external_benchmarks}

To evaluate whether the learned reasoning ability transfers beyond our task-progress benchmark, we additionally test the trained models on several public spatial and multi-image reasoning benchmarks. These experiments are included as supplementary evidence because they are not the primary optimization target of our method.

\begin{table*}[htbp]
  \centering
  \caption{OmniSpatial evaluation results.}
  \label{tab:omnispatial_results}
  {\setlength{\tabcolsep}{2pt}%
  \begin{tabularx}{\textwidth}{@{}l*{8}{>{\centering\arraybackslash}X}@{}}
    \toprule
    \textbf{Models} & \textbf{Manip.} & \textbf{Motion} & \textbf{Traffic} & \textbf{Locate} & \textbf{Geo.} & \textbf{Geom.} & \textbf{Ego} & \textbf{Hypo.} \\
    \midrule
    Random Choice & 24.86 & 26.30 & 25.88 & 23.43 & 27.27 & 24.77 & 22.55 & 25.78 \\
    GPT-3.5-turbo & 38.38 & 29.19 & 38.35 & 28.76 & 36.91 & 24.00 & 42.16 & 35.90 \\
    GPT-4-turbo & 42.97 & 37.40 & 41.18 & 28.95 & 40.00 & 26.32 & 31.37 & 35.42 \\
    STCR-Tag & 42.86 & 42.86 & 41.67 & 53.33 & 66.67 & 38.46 & 75.00 & 66.67 \\
    \bottomrule
  \end{tabularx}}
\end{table*}

\begin{table*}[htbp]
  \centering
  \caption{VSI-Bench evaluation results.}
  \label{tab:vsibench_results}
  {\setlength{\tabcolsep}{4pt}%
  \begin{tabularx}{\textwidth}{@{}l*{7}{>{\centering\arraybackslash}X}@{}}
    \toprule
    \textbf{Models} & \textbf{Avg.} & \textbf{Obj. Count} & \textbf{Abs. Dist.} & \textbf{Obj. Size} & \textbf{Rel. Dist.} & \textbf{Rel. Dir.} & \textbf{Appr. Order} \\
    \midrule
    LongVILA-8B & 21.6 & 29.1 & 9.1 & 16.7 & 29.6 & 30.7 & 25.5 \\
    InternVL2-2B & 26.5 & 25.7 & 24.0 & 20.0 & 32.1 & 44.1 & 6.3 \\
    VILA-1.5-8B & 28.9 & 17.4 & 21.8 & 50.3 & 32.1 & 34.8 & 24.8 \\
    STCR-Tag & 24.8 & 24.1 & 21.9 & 21.9 & 47.1 & 31.4 & 29.4 \\
    \bottomrule
  \end{tabularx}}
\end{table*}

\begin{table}[htbp]
  \centering
  \caption{CV-Bench evaluation results.}
  \label{tab:cvbench_results}
  \begin{tabularx}{\textwidth}{@{}l*{4}{>{\centering\arraybackslash}X}@{}}
    \toprule
    \textbf{Models} & \textbf{ADE20K} & \textbf{Omni3D} & \textbf{Depth} & \textbf{CV-Bench \cite{tong2024cambrian}} \\
    \midrule
    Qwen2.5-VL-7B & 68.56 & 75.17 & 70.83 & 75.07 \\
    STCR-CoT & 65.09 & 73.00 & 73.33 & 72.34 \\
    STCR-Tag & 67.30 & 76.42 & 79.17 & 75.27 \\
    \bottomrule
  \end{tabularx}
\end{table}

\begin{table*}[htbp]
  \centering
  \caption{MMSI-Bench evaluation results on compact categories.}
  \label{tab:mmsibench_compact_results}
  {\setlength{\tabcolsep}{3pt}%
  \begin{tabularx}{\textwidth}{@{}>{\raggedright\arraybackslash}p{0.24\textwidth}*{6}{>{\centering\arraybackslash}X}@{}}
    \toprule
    \textbf{Models} & \textbf{Cam.-Cam.} & \textbf{Obj.-Obj.} & \textbf{Reg.-Reg.} & \textbf{Cam.-Obj.} & \textbf{Obj.-Reg.} & \textbf{Cam.-Reg.} \\
    \midrule
    Qwen2.5-VL-7B & 24.7 & 24.5 & 24.7 & 25.6 & 29.4 & 26.5 \\
    STCR-CoT & 25.0 & 30.2 & 24.7 & 25.3 & 34.1 & 30.2 \\
    STCR-Tag & 27.9 & 33.0 & 32.1 & 33.7 & 36.5 & 32.1 \\
    \midrule
    \textbf{Models} & \textbf{Meas.} & \textbf{Appr.} & \textbf{Cam.} & \textbf{Obj.} & \textbf{MSR} & \textbf{Avg.} \\
    \midrule
    Qwen2.5-VL-7B & 25.0 & 18.2 & 20.3 & 39.5 & 25.8 & 25.9 \\
    STCR-CoT & 28.6 & 25.8 & 20.3 & 35.5 & 26.3 & 28.3 \\
    STCR-Tag & 29.7 & 24.2 & 16.2 & 34.2 & 27.3 & 29.6 \\
    \bottomrule
  \end{tabularx}}
\end{table*}

The results show that STCR-Tag transfers to several external spatial reasoning settings without direct optimization on those benchmarks. Gains are clearest on OmniSpatial \cite{jia2026omnispatial} and MMSI-Bench, while VSI-Bench \cite{yang2025thinking} remains mixed: the model improves relative-distance and appearance-order categories but does not outperform all baselines on the overall score. This suggests that our progressive training primarily strengthens relational spatial judgment and task-progress estimation, while general 3D scene measurement remains an open direction.

\section{Cross-Backbone Generalization}
\label{app:cross_backbone}

To test whether the training paradigm is tied to Qwen2.5-VL-7B, we additionally apply the full Stage 1 plus Stage 2 procedure to InternVL3-8B. Due to the limited turnaround time of the revision experiments, this reproduction is not as exhaustively tuned as the Qwen-based run. Nevertheless, it already yields consistent gains across all window sizes.

\begin{table*}[htbp]
  \centering
  \caption{Cross-backbone evaluation with InternVL3-8B under the progressive training pipeline.}
  \label{tab:internvl_full_results}
  {\setlength{\tabcolsep}{5pt}%
  \begin{tabularx}{\textwidth}{@{}l*{8}{>{\centering\arraybackslash}X}@{}}
    \toprule
    \textbf{Models} & \textbf{5} & \textbf{6} & \textbf{7} & \textbf{8} & \textbf{9} & \textbf{10} & \textbf{11} & \textbf{$\geq$12} \\
    \midrule
    InternVL3-8B & 47.7 & 50.6 & 49.4 & 47.9 & 47.1 & 48.7 & 48.9 & 51.0 \\
    InternVL3-8B-Full & 53.9 & 63.5 & 62.5 & 61.1 & 60.4 & 61.8 & 62.9 & 59.9 \\
    \bottomrule
  \end{tabularx}}
\end{table*}

These results indicate that the progressive training recipe is not specific to Qwen2.5-VL-7B. Even with a less optimized reproduction budget, InternVL3-8B benefits from the same CoT-to-tag progression, supporting the generality of the training paradigm across VLM backbones.

\section{Discussion}

Why did we choose to fine-tune Qwen2.5-VL?

We decided to choose Qwen2.5-VL-7B as the backbone model based on a thorough analysis of its untuned performance and key phenomena observed in subsequent fine-tuning experiments. First, Table~\ref{tab:model_evaluation} clearly shows Qwen2.5-VL-7B's exceptional performance in forward tasks without any fine-tuning. For instance, it achieved an average accuracy of up to 83.73\% across different window sizes, demonstrating its strong foundational capabilities. However, the table also revealed a significant performance drop in reverse tests, with an average accuracy of only 23.21\%. This indicates the model's bias but also highlights its substantial potential for improvement. Subsequently, we conducted LoRA fine-tuning experiments based on COT (Chain of Thought) on a small dataset. We observed a significant improvement in Qwen2.5-VL-7B's robustness and generalization ability, despite its initial bias. Table~\ref{tab:model_comparison} presents the specific manifestations and quantitative results of these performance enhancements. Collectively, these findings---Qwen2.5-VL-7B's strong forward capabilities in its untuned state and the significant optimization space and potential revealed through fine-tuning experiments---ultimately led us to select Qwen2.5-VL-7B as the foundational model for our research.

\begin{table*}[htbp]
  \centering
  \caption{Model accuracy on forward and reverse passes across different window sizes}
  \vspace{1em}
  \label{tab:model_evaluation}
  {\setlength{\tabcolsep}{2pt}
  \begin{tabularx}{\textwidth}{@{}>{\raggedright\arraybackslash}X*{9}{>{\centering\arraybackslash}p{0.060\textwidth}}@{}}
    \toprule
    \multirow{2}{*}{\textbf{Models}} & \multicolumn{9}{c}{\textbf{WinSize}}\\
    \cmidrule(lr){2-10}
    & \textbf{5} & \textbf{6} & \textbf{7} & \textbf{8} & \textbf{9} & \textbf{10} & \textbf{11} & \textbf{$\geq$12} & \textbf{Avg.}\\
    \midrule
    \rowcolor[HTML]{EFEFEF}
    \multicolumn{10}{l}{\textbf{2D Perception}} \\
    NVILA-8B (forward) & 30.3 & 43.4 & 36.3 & 50.5 & 43.4 & 44.4 & 40.4 & 46.4 & 41.89 \\
    NVILA-8B (reverse) & 62.6 & 60.6 & 55.6 & 61.6 & 56.6 & 55.6 & 54.6 & 63.6 & 58.85 \\
    Paligemma (forward) & 85.19 & 86.57 & 86 & 86.19 & 83.33 & 83.47 & 79.39 & 83.77 & 84.24 \\
    Paligemma (reverse) & 2.47 & 3.7 & 2.48 & 5.13 & 1.34 & 3.65 & 5.69 & 8.33 & 4.1 \\
    Qwen2.5-VL-7B (forward) & 83.8 & 78.8 & 82.5 & 82.1 & 83.8 & 87.1 & 86.3 & 85.4 & 83.73 \\
    Qwen2.5-VL-7B (reverse) & 23.8 & 22.5 & 24.7 & 22.5 & 23.3 & 21.7 & 27.1 & 20.1 & 23.21 \\
    \midrule
    \rowcolor[HTML]{EFEFEF}
    \multicolumn{10}{l}{\textbf{3D Perception}} \\
    3D-LLM (forward) & 29.3 & 33.7 & 35.2 & 33.6 & 37.9 & 36.8 & 40.2 & 39.7 & 35.8 \\
    3D-LLM (reverse) & 24.7 & 22.3 & 25.8 & 27.1 & 29.9 & 28.3 & 33.5 & 35.8 & 28.43 \\
    Chat-Scene (forward) & 41.4 & 43.4 & 51.3 & 40.7 & 38.7 & 42.8 & 46.6 & 58.1 & 45.38 \\
    Chat-Scene (reverse) & 20.3 & 21.6 & 30.1 & 57.4 & 44.5 & 47.9 & 40.1 & 43.9 & 38.23 \\
    3D-LLaVA (forward) & 38.4 & 44.4 & 44.4 & 46.3 & 48.8 & 48.8 & 47.9 & 47.3 & 45.79 \\
    3D-LLaVA (reverse) & 48 & 47.7 & 48.3 & 49.5 & 50.3 & 48.8 & 48 & 48.5 & 48.64 \\
    \bottomrule
  \end{tabularx}}%
\end{table*}

\begin{table*}[htbp]
    \centering
    \caption{Performance comparison of LoRA fine-tuned models across various window sizes.}
    \label{tab:model_comparison}
    \setlength{\tabcolsep}{4pt} 
    \renewcommand{\arraystretch}{1.2} 
    \begin{tabularx}{\textwidth}{l|*{8}{>{\centering\arraybackslash}X}}
        \toprule
        \textbf{Model\textbackslash WinSize} & \textbf{5} & \textbf{6} & \textbf{7} & \textbf{8} & \textbf{9} & \textbf{10} & \textbf{11} & \textbf{$\geq$12} \\
        \midrule
        \rowcolor[HTML]{EFEFEF}
        \multicolumn{9}{l}{\textbf{Baseline}} \\
        NVILA-8B (Forward) & 45.45 & 41.94 & 72.73 & 50.00 & 45.00 & 43.33 & 61.11 & 48.28 \\
        NVILA-8B (Backward) & 63.64 & 41.94 & 36.36 & 35.00 & 65.00 & 53.33 & 44.44 & 48.28 \\
        NVILA-8B (Average) & 54.55 & 41.94 & 54.55 & 42.50 & 55.00 & 48.33 & 52.78 & 48.28 \\
        Paligemma (Forward) & 77.78 & 94.12 & 90.91 & 66.67 & 92.86 & 88.24 & 100 & 69.23 \\
        Paligemma (Backward) & 76.92 & 85.71 & 69.23 & 66.67 & 100 & 68.75 & 88.89 & 72.22 \\
        Paligemma (Average) & 77.35 & 89.92 & 80.07 & 66.67 & 96.43 & 78.50 & 94.45 & 70.73 \\
        InternVL3-8B (Forward) & 44.44 & 47.06 & 63.64 & 33.33 & 69.23 & 87.50 & 44.44 & 58.33 \\
        InternVL3-8B (Backward) & 38.46 & 64.29 & 36.36 & 27.27 & 57.14 & 21.43 & 66.67 & 47.06 \\
        InternVL3-8B (Average) & 41.45 & 55.68 & 50.00 & 30.30 & 63.19 & 54.47 & 55.56 & 52.70 \\
        DeepSeek-VL2 (Forward) & 57.14 & 50.00 & 81.81 & 54.54 & 57.14 & 71.42 & 66.67 & 64.71 \\
        DeepSeek-VL2 (Backward) & 44.44 & 52.94 & 45.45 & 33.33 & 53.84 & 50.00 & 55.55 & 58.33 \\
        DeepSeek-VL2 (Average) & 50.79 & 51.47 & 63.63 & 43.94 & 55.49 & 60.71 & 61.11 & 61.52 \\
        LLaVA-OneVision-7B (Forward) & 61.54 & 57.14 & 45.45 & 54.54 & 42.86 & 57.14 & 55.56 & 70.59 \\
        LLaVA-OneVision-7B (Backward) & 22.22 & 35.29 & 18.18 & 66.67 & 53.84 & 62.50 & 44.44 & 28.57 \\
        LLaVA-OneVision-7B (Average) & 41.88 & 46.22 & 31.82 & 60.61 & 48.35 & 59.82 & 50.00 & 49.58 \\
        Qwen2.5-VL-7B (Forward) & 77.78 & 80.65 & 63.64 & 66.67 & 69.23 & 56.25 & 77.78 & 69.23 \\
        Qwen2.5-VL-7B (Backward) & 61.54 & 57.14 & 45.45 & 36.37 & 57.14 & 50.00 & 55.56 & 58.33 \\
        Qwen2.5-VL-7B (Average) & 69.66 & 68.90 & 54.55 & 51.52 & 63.19 & 53.13 & 66.67 & 63.78 \\
        \midrule
    \end{tabularx}
\end{table*}

\section{Potential Dense Reward Model and Safety Boundary for Imitation}
\subsection{Dense reward model}
In fact, by leveraging preference learning, our model can be used to supervise the real--time, online training of an embodied reward model. Unlike approaches that rely on sparse success signals or hand-crafted reward functions, our model provides dense, semantically-rich feedback online as the reinforcement learning (RL) policy interacts with the environment. Specifically, it assesses whether the policy's action proposal at each timestep contributes positively toward the task objective. By defining a visual reward function $r_{\phi}(\cdot)$ over a single-frame observation $o$, we can directly leverage pairwise preference supervision $y \in \{0, 1\}$ from a Vision-Language Model (VLM), indicating that one image is closer to task completion than another, to train a vision-based reward model via preference learning.

We model the robot policy rollout process as a Markov Decision Process (MDP) and use a neural network $r_{\phi}$, parameterized by $\phi$, to predict the reward $r_{\phi}(s_t, a_t)$ for the current state-action pair $o_t = (s_t, a_t)$. Adopting a Bradley-Terry style probabilistic model, we then convert the reward function $r_{\phi}(\cdot)$ into a probability of pairwise comparison between two observations:

\begin{equation}
\scalebox{0.9}{$
P_{\psi}[\tau^0 \succ \tau^1] =
\frac{\exp\!\left(\sum_{t \in \tau^0} r_{\psi}(o_t)\right)}
{\exp\!\left(\sum_{t \in \tau^0} r_{\psi}(o_t)\right) + \exp\!\left(\sum_{t \in \tau^1} r_{\psi}(o_t)\right)}
$}
\end{equation}

Periodically, every K timesteps, we feed the image pairs accumulated in the replay buffer into our Vision-Language Model (VLM) to obtain K preference labels, which are then stored in our preference buffer, D. The pairwise preference data in this buffer is subsequently used to supervise the reward model by minimizing the cross-entropy loss on the discrepancy between the preferences predicted by the reward model and those provided by the VLM.

\begin{equation} \label{eq:preference_loss}
\scalebox{0.7}{$
\mathcal{L}_{\text{pref}}(\psi) 
= -\mathbb{E}_{(\tau^0, \tau^1, y)} \Big[
    y \log P_{\psi}[\tau^0 \succ \tau^1] 
    + (1 - y) \log P_{\psi}[\tau^1 \succ \tau^0]
\Big]
$}
\end{equation}

This enables the online update of the reward model. Once the reward model $r_{\phi}(\cdot)$ is updated, it is used to re-calculate the instantaneous reward for every transition in the replay buffer. The robot's manipulation policy is then improved using these updated cumulative rewards via a standard RL loss, such as PPO \cite{schulman2017proximal}, facilitating the post-training of an RL policy from scratch or an imitation policy, such as a VLA. Through multi-round and multi-task policy rollouts, we ultimately obtain a general-purpose reward model whose preferences are aligned with our pretrained VLM. This model, in turn, helps the robot's manipulation policy to self-improve during its execution process.

\subsection{Safety boundary for imitation policy (VLA)}

A Vision-Language Model's (VLM) preference output on consecutive frames from a robot's manipulation process can provide a criterion for both safety protection and error recovery for imitation policies, such as the VLA. This gives open-loop imitation policies a real-time detection and recovery signal. Traditional imitation policies often struggle to recover from failure states during real-world deployment due to compounding errors, which can even lead to physical self-damage from large action errors. Intuitively, when a VLM provides a long sequence of negative preference feedback during a robot policy's rollout, it indicates that the policy is progressively deviating from the task objective or entering an out-of-control state. By setting a tolerable threshold for the maximum length of a cumulative negative preference sequence, the agent's erroneous execution can be terminated prematurely if this limit is exceeded. This simple criterion---triggering a policy interruption or reset---can reduce task failure from excessive cumulative error in long-horizon tasks and help detect potentially unsafe behavior during execution.

\section{More results}

Table~\ref{tab:evaluation_results} presents a quantitative analysis of how augmenting the weakly supervised tag dataset affects model performance. A clear observation from the results is a consistent improvement in performance across both forward and reverse evaluation metrics. This trend, where model efficacy scales with the volume of tag data, provides additional evidence for the scalability of our proposed paradigm.
\begin{table*}[htbp]
  \centering
  \caption{Evaluation results of varying Tag-to-CoT ratios across windows (Forward).}
  \label{tab:evaluation_results}
  {\setlength{\tabcolsep}{3pt}%
  \begin{tabularx}{\textwidth}{@{}l*{6}{>{\centering\arraybackslash}X}@{}}
    \toprule
    \textbf{Tag vs. CoT} & \textbf{5} & \textbf{6} & \textbf{7} & \textbf{8} & \textbf{9} & \textbf{10} \\
    \midrule
    14.3\% & 56.30\% & 55.96\% & 56.72\% & 56.33\% & 56.72\% & 56.90\% \\
    28.6\% & 28.49\% & 29.93\% & 31.34\% & 33.69\% & 34.93\% & 36.49\% \\
    42.9\% & 66.63\% & 70.15\% & 71.47\% & 73.48\% & 74.06\% & 74.95\% \\
    57.1\% & 67.17\% & 70.52\% & 71.68\% & 73.39\% & 74.08\% & 74.88\% \\
    71.4\% & 66.23\% & 69.70\% & 72.54\% & 74.50\% & 75.04\% & 77.34\% \\
    85.7\% & 68.68\% & 72.07\% & 74.13\% & 75.49\% & 77.04\% & 79.03\% \\
    100\% & 80.21\% & 82.81\% & 84.90\% & 85.99\% & 86.94\% & 87.44\% \\
    114.3\% & 78.45\% & 81.31\% & 83.28\% & 84.97\% & 85.71\% & 86.78\% \\
    128.6\% & 79.70\% & 81.98\% & 83.95\% & 84.48\% & 84.80\% & 86.13\% \\
    142.9\% & 79.06\% & 81.70\% & 83.19\% & 84.28\% & 84.80\% & 86.26\% \\
    157.1\% & 79.27\% & 81.82\% & 83.26\% & 83.87\% & 84.60\% & 85.35\% \\
    171.4\% & 77.28\% & 80.81\% & 82.64\% & 83.93\% & 84.66\% & 90.81\% \\
    \bottomrule
  \end{tabularx}}
  \vspace{0.6em}
  {\setlength{\tabcolsep}{3pt}%
  \begin{tabularx}{\textwidth}{@{}l*{7}{>{\centering\arraybackslash}X}@{}}
    \toprule
    \textbf{Tag vs. CoT} & \textbf{11} & \textbf{12} & \textbf{13} & \textbf{14} & \textbf{15} & \textbf{16} & \textbf{Overall} \\
    \midrule
    14.3\% & 57.10\% & 56.67\% & 55.99\% & 55.53\% & 57.76\% & 55.14\% & 56.32\% \\
    28.6\% & 37.68\% & 38.38\% & 38.56\% & 38.32\% & 37.88\% & 38.49\% & 34.63\% \\
    42.9\% & 74.59\% & 75.20\% & 75.55\% & 75.00\% & 75.00\% & 75.59\% & 72.99\% \\
    57.1\% & 76.09\% & 76.81\% & 76.43\% & 76.93\% & 77.13\% & 76.83\% & 73.72\% \\
    71.4\% & 78.75\% & 78.55\% & 78.77\% & 77.34\% & 78.75\% & 78.55\% & 75.04\% \\
    85.7\% & 79.41\% & 79.54\% & 80.55\% & 79.92\% & 81.06\% & 80.83\% & 76.54\% \\
    100\% & 88.09\% & 88.44\% & 88.87\% & 88.20\% & 88.87\% & 88.87\% & 86.04\% \\
    114.3\% & 87.34\% & 87.69\% & 88.01\% & 88.13\% & 88.32\% & 88.02\% & 85.05\% \\
    128.6\% & 86.60\% & 86.94\% & 87.53\% & 87.07\% & 87.77\% & 87.16\% & 84.06\% \\
    142.9\% & 86.10\% & 86.09\% & 86.27\% & 86.43\% & 86.31\% & 86.39\% & 84.31\% \\
    157.1\% & 85.90\% & 86.16\% & 86.16\% & 86.15\% & 86.18\% & 85.97\% & 84.15\% \\
    171.4\% & 85.63\% & 86.30\% & 86.27\% & 86.27\% & 86.18\% & 85.92\% & 83.80\% \\
    \bottomrule
  \end{tabularx}}
\end{table*}

\begin{table*}[htbp]
  \ContinuedFloat
  \centering
  \caption{Evaluation results of varying Tag-to-CoT ratios across windows (Reverse).}
  {\setlength{\tabcolsep}{3pt}%
  \begin{tabularx}{\textwidth}{@{}l*{6}{>{\centering\arraybackslash}X}@{}}
    \toprule
    \textbf{Tag vs. CoT} & \textbf{5} & \textbf{6} & \textbf{7} & \textbf{8} & \textbf{9} & \textbf{10} \\
    \midrule
    14.3\% & 51.72\% & 52.07\% & 53.98\% & 54.73\% & 54.17\% & 55.47\% \\
    28.6\% & 92.73\% & 93.18\% & 93.65\% & 94.35\% & 93.91\% & 94.73\% \\
    42.9\% & 74.27\% & 75.91\% & 78.16\% & 79.07\% & 80.53\% & 80.65\% \\
    57.1\% & 76.82\% & 79.63\% & 81.23\% & 82.58\% & 84.28\% & 85.29\% \\
    71.4\% & 82.52\% & 83.77\% & 84.80\% & 85.61\% & 86.35\% & 87.35\% \\
    85.7\% & 82.61\% & 84.53\% & 85.87\% & 86.74\% & 87.38\% & 88.16\% \\
    100\% & 75.25\% & 77.17\% & 78.66\% & 79.68\% & 81.49\% & 81.86\% \\
    114.3\% & 78.50\% & 80.43\% & 81.84\% & 82.94\% & 83.99\% & 84.49\% \\
    128.6\% & 81.75\% & 83.69\% & 85.01\% & 86.19\% & 86.50\% & 87.13\% \\
    142.9\% & 82.87\% & 85.05\% & 85.81\% & 87.07\% & 86.97\% & 87.82\% \\
    157.1\% & 82.26\% & 84.43\% & 85.61\% & 86.71\% & 87.67\% & 88.41\% \\
    171.4\% & 87.47\% & 88.79\% & 89.85\% & 90.35\% & 85.73\% & 91.43\% \\
    \bottomrule
  \end{tabularx}}
  \vspace{0.6em}
  {\setlength{\tabcolsep}{3pt}%
  \begin{tabularx}{\textwidth}{@{}l*{7}{>{\centering\arraybackslash}X}@{}}
    \toprule
    \textbf{Tag vs. CoT} & \textbf{11} & \textbf{12} & \textbf{13} & \textbf{14} & \textbf{15} & \textbf{16} & \textbf{Overall} \\
    \midrule
    14.3\% & 56.53\% & 56.92\% & 57.63\% & 57.21\% & 59.16\% & 59.37\% & 55.23\% \\
    28.6\% & 94.71\% & 94.52\% & 94.62\% & 95.08\% & 95.10\% & 95.05\% & 94.15\% \\
    42.9\% & 81.38\% & 81.54\% & 81.88\% & 81.89\% & 81.63\% & 81.68\% & 79.40\% \\
    57.1\% & 85.27\% & 86.27\% & 86.92\% & 86.89\% & 86.71\% & 87.01\% & 83.42\% \\
    71.4\% & 87.76\% & 88.01\% & 87.87\% & 87.83\% & 87.64\% & 87.73\% & 86.09\% \\
    85.7\% & 88.89\% & 88.51\% & 89.21\% & 89.14\% & 88.83\% & 89.20\% & 87.02\% \\
    100\% & 82.34\% & 82.50\% & 82.50\% & 82.75\% & 82.30\% & 82.30\% & 80.28\% \\
    114.3\% & 84.77\% & 84.97\% & 84.91\% & 85.20\% & 84.44\% & 84.58\% & 83.04\% \\
    128.6\% & 87.20\% & 87.44\% & 87.82\% & 87.66\% & 87.88\% & 87.57\% & 86.69\% \\
    142.9\% & 87.96\% & 88.22\% & 88.22\% & 87.97\% & 88.22\% & 88.22\% & 87.13\% \\
    157.1\% & 88.96\% & 89.79\% & 89.17\% & 89.67\% & 89.45\% & 88.82\% & 87.27\% \\
    171.4\% & 91.55\% & 91.85\% & 91.42\% & 91.23\% & 90.99\% & 90.68\% & 90.33\% \\
    \bottomrule
  \end{tabularx}}
\end{table*}

\begin{table*}[htbp]
  \ContinuedFloat
  \centering
  \caption{Evaluation results of varying Tag-to-CoT ratios across windows (Average).}
  {\setlength{\tabcolsep}{3pt}%
  \begin{tabularx}{\textwidth}{@{}l*{6}{>{\centering\arraybackslash}X}@{}}
    \toprule
    \textbf{Tag vs. CoT} & \textbf{5} & \textbf{6} & \textbf{7} & \textbf{8} & \textbf{9} & \textbf{10} \\
    \midrule
    14.3\% & 54.01\% & 54.01\% & 55.35\% & 55.53\% & 55.45\% & 56.19\% \\
    28.6\% & 60.61\% & 61.55\% & 62.49\% & 64.02\% & 64.42\% & 65.61\% \\
    42.9\% & 70.45\% & 73.03\% & 74.82\% & 76.28\% & 77.29\% & 77.80\% \\
    57.1\% & 71.99\% & 75.07\% & 76.46\% & 77.98\% & 79.18\% & 80.09\% \\
    71.4\% & 74.37\% & 76.74\% & 78.67\% & 80.05\% & 80.97\% & 82.35\% \\
    85.7\% & 75.65\% & 78.30\% & 80.00\% & 81.11\% & 82.21\% & 83.59\% \\
    100\% & 77.73\% & 79.99\% & 81.78\% & 82.84\% & 84.22\% & 85.65\% \\
    114.3\% & 78.48\% & 80.87\% & 82.56\% & 83.96\% & 84.85\% & 85.64\% \\
    128.6\% & 79.22\% & 81.75\% & 83.43\% & 85.07\% & 86.50\% & 86.63\% \\
    142.9\% & 80.97\% & 83.37\% & 84.50\% & 85.68\% & 85.89\% & 87.04\% \\
    157.1\% & 80.77\% & 83.13\% & 84.44\% & 85.29\% & 86.13\% & 86.88\% \\
    171.4\% & 82.38\% & 84.80\% & 86.24\% & 87.14\% & 87.73\% & 88.58\% \\
    \bottomrule
  \end{tabularx}}
  \vspace{0.6em}
  {\setlength{\tabcolsep}{3pt}%
  \begin{tabularx}{\textwidth}{@{}l*{7}{>{\centering\arraybackslash}X}@{}}
    \toprule
    \textbf{Tag vs. CoT} & \textbf{11} & \textbf{12} & \textbf{13} & \textbf{14} & \textbf{15} & \textbf{16} & \textbf{Overall} \\
    \midrule
    14.3\% & 56.82\% & 56.79\% & 56.81\% & 56.37\% & 57.46\% & 57.25\% & 55.78\% \\
    28.6\% & 66.20\% & 66.45\% & 66.59\% & 66.70\% & 66.49\% & 66.77\% & 64.39\% \\
    42.9\% & 77.98\% & 78.37\% & 78.72\% & 78.44\% & 78.34\% & 78.64\% & 76.20\% \\
    57.1\% & 80.68\% & 81.54\% & 81.67\% & 81.91\% & 81.92\% & 81.92\% & 78.57\% \\
    71.4\% & 83.26\% & 83.28\% & 83.60\% & 83.71\% & 83.53\% & 83.35\% & 80.56\% \\
    85.7\% & 84.15\% & 84.03\% & 84.88\% & 84.53\% & 84.94\% & 85.01\% & 81.78\% \\
    100\% & 85.22\% & 85.47\% & 85.83\% & 85.47\% & 85.14\% & 85.59\% & 83.16\% \\
    114.3\% & 86.06\% & 86.33\% & 86.46\% & 86.67\% & 86.38\% & 86.30\% & 84.05\% \\
    128.6\% & 86.90\% & 87.19\% & 87.40\% & 87.87\% & 87.17\% & 87.01\% & 84.93\% \\
    142.9\% & 87.43\% & 87.16\% & 87.24\% & 87.38\% & 88.06\% & 86.99\% & 85.50\% \\
    157.1\% & 87.43\% & 87.98\% & 87.66\% & 87.91\% & 87.91\% & 87.39\% & 85.64\% \\
    171.4\% & 88.59\% & 89.08\% & 88.84\% & 88.75\% & 88.59\% & 88.30\% & 87.07\% \\
    \bottomrule
  \end{tabularx}}
\end{table*}

\begin{table*}[htbp]
\centering
\caption{MMSI-Bench evaluation results (positional relationships).}
\label{tab:mmsibench_results}
{\setlength{\tabcolsep}{3pt}%
\begin{tabularx}{\textwidth}{@{}>{\raggedright\arraybackslash}p{0.24\textwidth}*{6}{>{\centering\arraybackslash}X}@{}}
\toprule
\textbf{Models} & \textbf{Cam.-Cam.} & \textbf{Obj.-Obj.} & \textbf{Reg.-Reg.} & \textbf{Cam.-Obj.} & \textbf{Obj.-Reg.} & \textbf{Cam.-Reg.} \\
\midrule
\rowcolor[HTML]{EFEFEF}
\multicolumn{7}{l}{\textbf{Baseline}} \\
InternVL3-78B & 34.4 & 23.4 & 32.1 & 12.8 & 37.6 & 26.5 \\
InternVL2.5-78B & 23.7 & 22.3 & \textbf{39.5} & 29.1 & 31.8 & \textbf{42.2} \\
LLaVA-OneVision-72B & \textbf{43.0} & 31.9 & 33.3 & 30.2 & 37.6 & 38.6 \\
InternVL2.5-38B & 18.3 & 22.3 & 35.8 & 22.1 & \textbf{38.8} & 34.9 \\
Qwen2.5-VL-32B & 24.7 & 26.6 & 29.6 & 22.1 & 32.9 & 31.3 \\
InternVL2.5-26B & 24.7 & 19.1 & 29.6 & 33.7 & 31.8 & 37.3 \\
InternVL3-14B & 19.4 & 24.5 & 24.7 & 23.3 & 37.6 & 24.1 \\
Llama-3.2-11B-Vision & 25.8 & 30.8 & 32.0 & 25.6 & 21.2 & 25.9 \\
InternVL3-9B & 18.3 & 25.5 & 32.1 & 29.1 & 31.8 & 22.9 \\
InternVL3-8B & 25.8 & 31.9 & 37.0 & 25.6 & 35.3 & 28.9 \\
InternVL2.5-8B & 32.3 & 27.7 & 29.6 & 32.6 & 24.7 & 32.5 \\
NVILA-8B & 17.2 & 29.8 & 24.7 & 30.2 & 22.4 & 34.9 \\
\rowcolor[HTML]{EFEFEF}
Qwen2.5-VL-7B & 24.7 & 24.5 & 24.7 & 25.6 & 29.4 & 26.5 \\
LLaVA-OneVision-7B & 20.4 & \textbf{33.0} & 29.6 & 29.1 & 25.9 & 30.1 \\
InternVL2.5-4B & 31.2 & 23.4 & 21.0 & 31.4 & 34.1 & 25.3 \\
Qwen2.5-VL-3B & 26.9 & 27.7 & 30.9 & 29.1 & 28.2 & 34.9 \\
InternVL3-2B & 26.9 & 25.5 & 29.6 & 31.4 & 28.2 & 27.7 \\
InternVL2.5-2B & 28.0 & 27.7 & 24.7 & \textbf{37.2} & 29.4 & 36.1 \\
InternVL3-1B & 24.7 & 35.1 & 22.2 & 30.2 & 29.4 & 30.1 \\
InternVL2.5-1B & 23.7 & 26.6 & 24.7 & 25.6 & 31.8 & 25.3 \\
DeepSeek-VL2 & 23.7 & 31.9 & 22.2 & 36.0 & 30.6 & 22.9 \\
DeepSeek-VL2-Small & 24.7 & 28.7 & 18.5 & 33.7 & \textbf{38.8} & 27.7 \\
DeepSeek-VL2-Tiny & 29.0 & 27.7 & 21.0 & 23.3 & 17.6 & 31.3 \\
\midrule
\rowcolor[HTML]{EFEFEF}
\multicolumn{7}{l}{\textbf{Our models}} \\
STCR-CoT & 25.0 & 30.2 & 24.7 & 25.3 & 34.1 & 30.2 \\
STCR-Tag & 27.9 & 33.0 & 32.1 & 33.7 & 36.5 & 32.1 \\
\bottomrule
\end{tabularx}}
\end{table*}

\begin{table*}[htbp]
\ContinuedFloat
\centering
\caption{MMSI-Bench evaluation results (attribute, motion, and summary).}
{\setlength{\tabcolsep}{3pt}%
\begin{tabularx}{\textwidth}{@{}>{\raggedright\arraybackslash}p{0.24\textwidth}*{6}{>{\centering\arraybackslash}X}@{}}
\toprule
\textbf{Models} & \textbf{Meas.} & \textbf{Appr.} & \textbf{Cam.} & \textbf{Obj.} & \textbf{MSR} & \textbf{Avg.} \\
\midrule
\rowcolor[HTML]{EFEFEF}
\multicolumn{7}{l}{\textbf{Baseline}} \\
InternVL3-78B & 37.5 & 19.7 & \textbf{28.4} & 31.6 & 29.3 & 28.5 \\
InternVL2.5-78B & 35.9 & 19.7 & 17.6 & 26.3 & 27.3 & 28.5 \\
LLaVA-OneVision-72B & 28.1 & 19.7 & 13.5 & 32.9 & 15.7 & 28.4 \\
InternVL2.5-38B & 37.5 & 25.8 & 14.9 & 38.2 & 25.3 & 27.9 \\
Qwen2.5-VL-32B & 31.2 & 24.2 & 18.9 & 35.5 & 27.8 & 27.7 \\
InternVL2.5-26B & 35.9 & 30.3 & 10.8 & 31.6 & 26.8 & 28.0 \\
InternVL3-14B & 31.2 & 22.7 & 24.3 & 31.6 & 29.3 & 26.8 \\
Llama-3.2-11B-Vision & 20.3 & 19.7 & 25.6 & 28.9 & 19.2 & 25.4 \\
InternVL3-9B & 29.7 & 24.2 & 16.2 & 38.2 & 26.8 & 26.7 \\
InternVL3-8B & 23.4 & 24.2 & 16.2 & 32.9 & 14.6 & 25.7 \\
InternVL2.5-8B & 26.6 & 27.3 & 16.2 & 31.6 & \textbf{30.3} & 28.7 \\
NVILA-8B & 34.4 & 25.8 & 25.7 & 34.2 & 29.8 & 28.1 \\
\rowcolor[HTML]{EFEFEF}
Qwen2.5-VL-7B & 25.0 & 18.2 & 20.3 & \textbf{39.5} & 25.8 & 25.9 \\
LLaVA-OneVision-7B & 29.7 & 25.8 & 18.9 & 34.2 & 11.6 & 24.5 \\
InternVL2.5-4B & 23.4 & 24.2 & 13.5 & 31.6 & 26.8 & 26.3 \\
Qwen2.5-VL-3B & 31.2 & 16.7 & 17.6 & 27.6 & 23.2 & 26.5 \\
InternVL3-2B & 26.6 & 22.7 & 12.2 & 23.7 & 23.7 & 25.3 \\
InternVL2.5-2B & \textbf{43.8} & 15.2 & 21.6 & 31.6 & 26.8 & \textbf{29.0} \\
InternVL3-1B & 32.8 & 28.8 & 17.6 & 19.7 & 26.3 & 27.0 \\
InternVL2.5-1B & 31.2 & 30.3 & 17.6 & 25.0 & 26.3 & 26.1 \\
DeepSeek-VL2 & 28.1 & 15.2 & \textbf{28.4} & 26.3 & 28.3 & 27.1 \\
DeepSeek-VL2-Small & 28.1 & \textbf{33.3} & 24.3 & 25.0 & 29.8 & 28.6 \\
DeepSeek-VL2-Tiny & 14.1 & 24.2 & 14.9 & 25.0 & 27.3 & 24.0 \\
\midrule
\rowcolor[HTML]{EFEFEF}
\multicolumn{7}{l}{\textbf{Our models}} \\
STCR-CoT & 28.6 & 25.8 & 20.3 & 35.5 & 26.3 & 28.3 \\
STCR-Tag & 29.7 & 24.2 & 16.2 & 34.2 & 27.3 & 29.6 \\
\bottomrule
\end{tabularx}}
\end{table*}

As shown in Table ~\ref{tab:mmsibench_results}, the model, after being trained on Qwen2.5-VL-7B, demonstrates improved performance on the MMSI-Bench \cite{yang2025mmsi} benchmark. 
Following CoT training, our model (CoT) shows an increase in average accuracy (Avg.) from 25.9\% to 28.3\%, which is further boosted to 29.6\% after weakly supervised fine-tuning with Tag. 
This result gives STCR-Tag the best average score among the tested models in Table~\ref{tab:mmsibench_results}.
Furthermore, we observe that the performance gains are concentrated in the key dimensions of spatial perception and reasoning. 
Specifically, appearance attribute perception (Appr.) increased from 18.2\% to 24.2\%, indicating an enhanced fine-grained recognition of intrinsic object features. 
The accuracy of reasoning about object-object relationships improved from 24.5\% to 33.0\%, an 7.1\% increase, showing a strengthened ability to understand and judge the relative spatial relationships between multiple objects. 
Moreover, reasoning about object-region relationships rose from 29.4\% to 36.5\%, a 7.1\% improvement, demonstrating a more precise understanding of object localization and relationships within spatial areas.
\begin{figure*}[t]
    \centering
    \includegraphics[width=\textwidth]{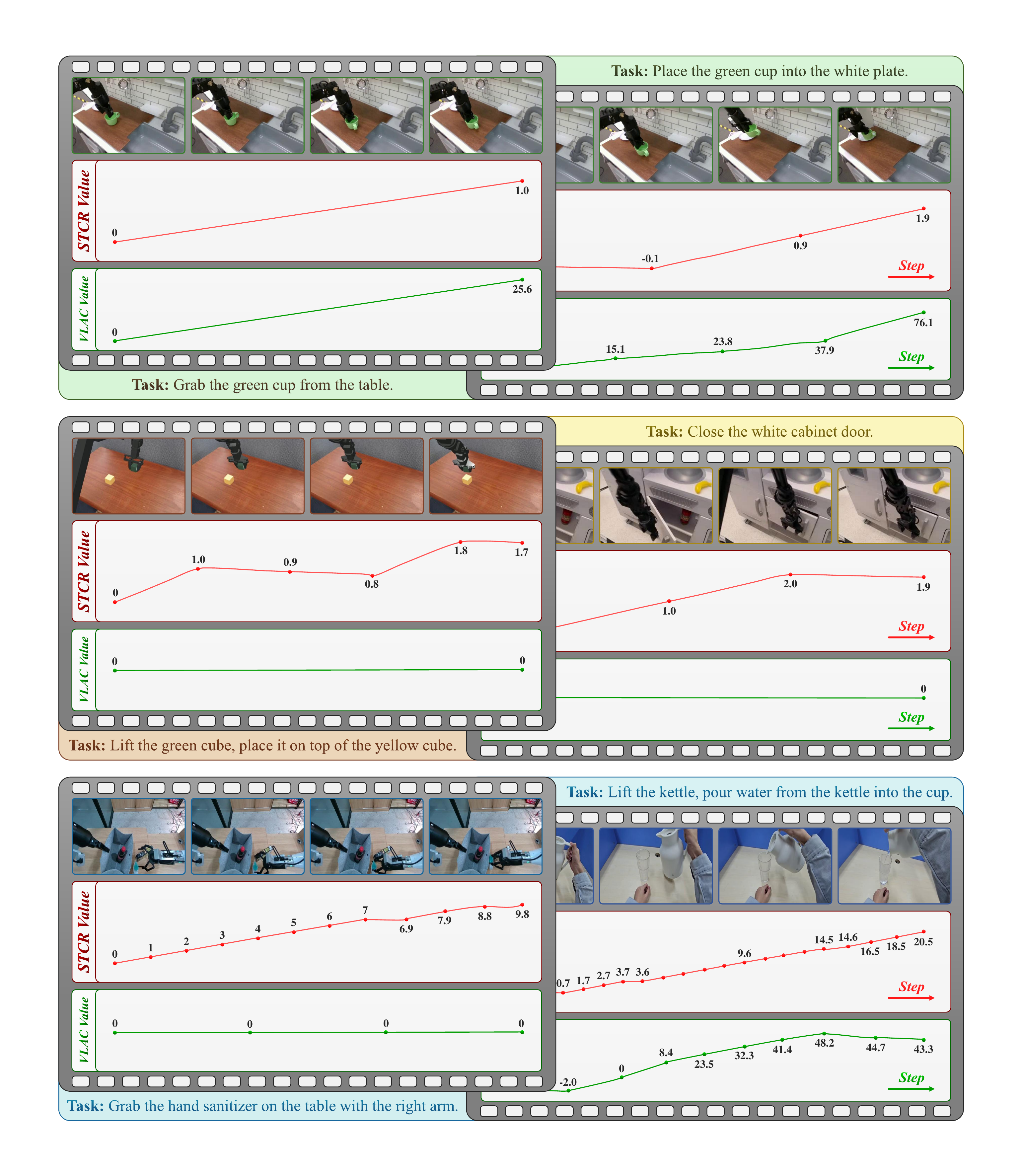} 
    \caption{
        \textbf{ Comparison with VLAC.} 
    }
    \label{fig:comparison_with_vlac}
\end{figure*}

\end{document}